\documentclass{article}

 \usepackage[nonatbib, dblblindworkshop, final]{neurips_2025}

\usepackage[utf8]{inputenc} 
\usepackage[T1]{fontenc}    
\usepackage{hyperref}       
\usepackage{url}            
\usepackage{booktabs}       
\usepackage{amsfonts}       
\usepackage{nicefrac}       
\usepackage{microtype}      
\usepackage{xcolor}         

\usepackage{float}
\usepackage{graphicx}
\usepackage{amsmath}
\usepackage{amssymb}
\usepackage{amsthm}
\usepackage{mathtools}
\usepackage{booktabs}
\usepackage{tabularx}
\usepackage{multirow}
\usepackage{siunitx}    
\usepackage{multirow}   
\usepackage{geometry}
\usepackage{xcolor}
\usepackage{pifont}
\usepackage{booktabs,longtable,siunitx}
\usepackage{amsthm}

\newtheorem{theorem}{Theorem}[section]

\theoremstyle{definition}
\newtheorem{definition}[theorem]{Definition}

\theoremstyle{remark}

\usepackage[ruled,vlined,linesnumbered]{algorithm2e}
\SetKwComment{Comment}{$\triangleright$\ }{}
\SetKwFunction{FNestedCV}{NestedCVEvaluation}
\SetKwFunction{FFinalEval}{FinalModelEvaluation}
\SetKwFunction{FHPO}{FindBestHyperparameters}
\SetKwFunction{FTrain}{TrainModel}
\SetKwFunction{FEval}{Evaluate}
\SetKwFunction{FBootstrap}{BootstrapCI}
\SetKwInOut{KwIn}{Require}
\SetKwInOut{KwOut}{Ensure}

\newcommand{\cmark}{\textcolor{green!70!black}{\ding{51}}}

\newcommand{\gminus}{\textcolor{gray}{\rule[0.5ex]{0.8em}{0.55pt}}}

\usepackage[style=numeric, citestyle=authoryear-comp, maxcitenames=2, uniquename=false]{biblatex}
\let\citep=\parencite
\let\citet=\textcite

\title{Topological Feature Compression for Molecular Graph Neural Networks}
\workshoptitle{AI4Science}

\author{%
  Rahul Khorana \\
  Department of Computing\\
  Imperial College London\\
  London, SW7 2AZ, UK \\
  \texttt{rahul.khorana24@imperial.ac.uk}
}

\begin{document}

\maketitle

\begin{abstract}
Recent advances in molecular representation learning have produced highly effective encodings of molecules for numerous cheminformatics and bioinformatics tasks. However, extracting general chemical insight while balancing predictive accuracy, interpretability, and computational efficiency remains a major challenge. In this work, we introduce a novel Graph Neural Network (GNN) architecture that combines compressed higher-order topological signals with standard molecular features. Our approach captures global geometric information while preserving computational tractability and human-interpretable structure. We evaluate our model across a range of benchmarks, from small-molecule datasets to complex material datasets, and demonstrate superior performance using a parameter-efficient architecture. We achieve the best performing results in both accuracy and robustness across almost all benchmarks. We open source all code \footnote{All code and results can be found on Github \url{https://github.com/rahulkhorana/TFC-PACT-Net}}.
\end{abstract}

\section{Introduction}


Machine learning has emerged as a powerful paradigm for modeling biochemical systems across various levels of chemical scales \citep{wang2019molecule, zitnik2023a, wang2022molecular, wang2021a, wang2023scientific, somnath2021multi}. A predominant approach relies on one-dimensional, string-based representations such as SMILES and SELFIES, or derived fingerprints like ECFP. However, these representations fundamentally lack explicit encoding of 3D geometry and topology. This inherent limitation curtails their expressivity for structure-sensitive tasks crucial to drug discovery and materials science, including quantitative structure-activity relationship (QSAR) analysis, molecular docking, de novo drug design, compound identification, and molecular property prediction \citep{guha2013exploring, mckinney2000practice, ma2011molecular, williams2008public, hartenfeller2010novo, keith2021combining, khorana2024polyatomiccomplexestopologicallyinformedlearning, esders2025analyzing}. 

To imbue models with spatial and relational inductive biases, recent efforts have focused on geometrically-aware architectures, principally Graph Neural Networks (GNNs) and, more recently, higher-order structures like simplicial and cellular complexes \citep{bodnar2021weisfeiler, bodnar2023topological, goh2022simplicial, wu2023simplicial, khorana2024polyatomiccomplexestopologicallyinformedlearning, verma2024topological}. Despite their enhanced representational power, these methods often present a challenging trade-off between model complexity and predictive performance. Paradoxically, on certain benchmark tasks, their performance, as measured by standard regression metrics (RMSE/MAE), can be surpassed by simpler, non-geometric baselines, highlighting issues of scalability, optimization, and generalization.

At the other end of the spectrum, methods grounded in first principles, such as Density Functional Theory (DFT), and surrogate Machine Learning Interaction Potentials (MLIPs) offer high-fidelity predictions by approximating quantum mechanical interactions \citep{butera2024density, huang2023central, friederich2021machine}. Nonetheless, their substantial computational cost, often scaling poorly with system size, renders them computationally infeasible. This establishes a clear trade-off in the field: a choice between computationally efficient but geometrically impoverished representations, geometrically aware yet poorly performing representations, and highly accurate but computationally prohibitive first-principles calculations. 
To address this challenge, graph neural networks (GNNs) have become a widely adopted tool \citep{GAT, GCN, GIN, SAGE, wang2022molecular, zang2023hierarchical}. However, the conventional approach of using molecular graphs as input for GNNs may inadvertently constrain their potential for generating robust representations and performing accurate property prediction \citep{jiang2025bi}.

In this study, we introduce \textsc{PACTNet}, a new graph neural network specifically designed to leverage compressed higher-order topological features from cellular representations. We create these enriched knowledge graphs fusing these compressed features with knowledge graphs using our efficient cellular compression (\textsc{ECC}) algorithm. Our approach enables us to capture features representative of complex 3D molecular structures, producing robust embeddings, leading to improved property prediction performance on numerous benchmarks across many levels of chemical scale, from small molecules to complex biomolecules (protein-ligand complexes), and quantum properties. We demonstrate the empirically best performing approach in RMSE and MAE across all but two datasets. 
Our major contributions can be summarized as follows:
\begin{itemize}
    \item \textbf{Novel Topological Feature Integration.} We introduce \textsc{ECC}, a method for augmenting molecular graphs with compressed features derived from higher-order cellular complexes. This technique creates a topologically-informed graph representation that enriches the node and edge features with multi-dimensional geometric and relational data. We demonstrate that this method provides a principled way to distill complex topological information into a standard graph structure, enhancing downstream model performance without requiring specialized higher-order architectures.
    \item \textbf{Computationally Efficient, Geometrically-Aware Molecular Embeddings.} We propose a new representation learning framework that resolves the prevailing trade-off between geometric fidelity and computational cost. By leveraging features from cellular complexes, our method generates embeddings that (1) retain crucial 3D structural information, overcoming the limitations of string representations; (2) demonstrate superior performance and robustness over standard GNNs across a diverse set of chemical tasks and scales; and (3) remain orders of magnitude more computationally efficient than first-principles methods like DFT.
    \item \textbf{A Novel Graph Neural Network.} We propose the \textsc{PACTNet}, a GNN that synergistically combines three distinct classes of features: (1) local neighborhood structure via principal neighborhood aggregation, (2) global higher-order topology via spectral features, and (3) node-level connectivity statistics via degree histograms. This multi-faceted aggregation scheme allows our resulting network, \textsc{PACTNet}, to capture a richer hierarchy of structural information than prior methods, demonstrably improving its expressive power and imbuing it with strong, chemically-relevant inductive biases.
\end{itemize}

To empirically validate our proposed methods, we conducted a comprehensive evaluation of \textsc{PACTNet} on seven benchmark datasets for molecular property prediction, encompassing diverse chemical properties and molecular scales. The performance of our model was benchmarked against a suite of well-established GNN architectures, including GCN \citep{GCN}, GAT \citep{GAT}, GraphSAGE \citep{SAGE}, and GIN \citep{GIN}. GIN is considered to be a gold standard benchmark in general graph based molecular learning \citep{fey2020hierarchical}. GCN and GAT are similarly considered to be competitive baseline models for graph based molecular learning \citep{jiang2021could}.

Our results demonstrate that \textsc{PACTNet} establishes a new, high-performing baseline, significantly outperforming these widely-adopted models on five of the seven tasks with an empirical reduction in Root Mean Squared Error (RMSE).

\section{Background \& Related Work}

\subsection{Molecular Representations \& Embeddings}

Molecular machine learning faces a persistent trade-off between geometric fidelity, predictive accuracy, and computational cost. String-based encodings such as SMILES \citep{SMILES}, SELFIES \citep{SELFIES}, and fingerprint derivatives like ECFP \citep{ECFP} are efficient but discard 3D structural information critical to molecular function \citep{liu2021pre}. Geometrically aware methods, ranging from 3D GNNs \citep{alhamoud2024leveraging, godwin2021simple} to polyatomic complexes \citep{khorana2024polyatomiccomplexestopologicallyinformedlearning}, offer richer representations but often fail to surpass simpler baselines. At the other extreme, first-principles models such as MLIPs and Behler–Parrinello networks \citep{lysogorskiy2021performant, dusson2022atomic, zuo2020performance} achieve high accuracy yet remain computationally prohibitive for large-scale screening.

Our framework seeks a middle ground, combining the geometric expressivity of topological methods with the predictive power of learned embeddings, while retaining practical scalability.

\subsection{Graph Neural Networks}

Operating directly on 2D molecular graphs, GNNs such as GCN \citep{GCN}, GraphSAGE \citep{SAGE}, GAT \citep{GAT}, and GIN \citep{GIN} perform message passing, iteratively aggregating local neighborhoods \citep{zhong2023hierarchical}. This approach is limited by the Weisfeiler–Leman graph isomorphism test and struggles to capture long-range or global topological structure without deep, unstable architectures \citep{feng2022powerful, balcilar2021breaking}. Even specialized 3D models like PointNet++ \citep{qi2017pointnet}, SchNet \citep{schutt2018schnet}, and ForceNet \citep{hu2021forcenet} address some of these issues but incur greater computational cost.

We address these limitations by enriching message passing with higher-order topological features, giving GNNs direct access to geometric and global information they cannot efficiently learn alone, improving robustness and expressivity without resorting to full 3D or first-principles methods.

\section{Methods}

In this section we present our PactNet neural network, the PACT-linear Layer (\textsc{PACTLayer}), and ECC representation. We define the following key concepts below.

\begin{definition} (\textbf{Molecular Graph}) A molecular graph (MG) is a structured representation of a molecule, namely a graph $\mathcal{G} = (V,E)$ where $V$ is the vertex set and $E$ the edge set. In the case of molecular graphs, $V$ contains the atoms and $E$ contains all bonds. Therefore providing a structured way in which to represent a molecule.
\end{definition}

\begin{definition} (\textbf{Knowledge Graph}) A knowledge graph (KG) is a structured representation of knowledge in which nodes are connected by relations or edges. Formally a directed knowledge graph is represented as a set of triples $\mathcal{T} = \{(h,r,t)_i\}_{i=1}^{n}$ where each triple contains a head entity $h$, tail entity $t$, and relation $r$ connecting them. A knowledge graph can also be viewed as a graph $\mathcal{G} = (V,E)$.
\end{definition}

\begin{definition} (\textbf{Lifting Transformation}, 
\label{def:lifting-tform}
\citet{bodnar2021weisfeiler})  A cellular lifting transformation is a function $f : \mathcal{G} \to \mathcal{X}$ from the space of graphs $\mathcal{G}$ to the space of regular cell complexes $\mathcal{X}$ with the property that two graphs $\mathcal{G}_1$, $\mathcal{G}_2$ are isomorphic iff the cell complexes $f(\mathcal{G}_1)$, $f(\mathcal{G}_2)$ are isomorphic.
\end{definition}

\subsection{Topological Compression \& ECC Algorithm}

\paragraph{The need for compression} Formally, we want to leverage Cell Complexes to provide higher order geometric information to our model. In areas such as materials design, quantum chemistry, and protein informatics the geometry of the molecule is of fundamental importance for property prediction and design \citep{wang2024enhancing, krapp2024context}. However, cell complexes can be high dimensional and complex to learn over \citep{bodnar2021weisfeiler}. In order to reduce the complexity, we compress relevant information that can be extracted from the cell complex.

\paragraph{ECC Algorithm} Given a molecular graph $G$ we use a Molecular-Lifting Transformation, which is a kind of lifting transformation as in definition \ref{def:lifting-tform}. 

\begin{definition} (\textbf{Molecular-Lifting Transformation})
Formally, let $M$ be a molecule and $G_{M}$ be the corresponding molecular graph. Then we define a function $\mathcal{F}$ which sends $G_M \mapsto X_M$, where $X_M$ is a cell complex. In particular, given a set of atoms in the molecule contained in $V_M := \{A_1, \ldots, A_n\}$ we determine information about the number of protons, neutrons and electrons for each atom, and use these as base-points $X^{0}_{M}$ ($0$-cells). We encapsulate these inside the individual atoms via attachment ($1$-cells). Then we connect the bonds ($2$-cells). Finally we consider induced cycles, chemical rings, and $k$-hop interactions ($3$-cells). Thus we end up with, for molecule $M$, the corresponding $\{X^{0}_M, X^{1}_M, X^2_M, X^3_M\}$. The resulting cell complex is termed $X_M$ and skeleton preserving (isomorphic in the sense that $f(X^2_M)$ corresponds to the molecular graph $G$).
\end{definition}

This type of molecular lifting map is more complex than the scheme proposed by \citet{bodnar2021weisfeiler}, yet less complex and containing far less information than the representation developed by \citet{khorana2024polyatomiccomplexestopologicallyinformedlearning}. The map $\mathcal{F}$ occupies an optimal middle ground, balancing representational complexity, geometric information, and computational cost.

Therefore, upon transforming molecule $M$ to cell complex $\mathcal{F}(M)$, we can easily extract relevant topologically rich features. We compute the betti-numbers, take the eigen-decomposition of the chain matrix (termed spectral chains), accumulate the top-$k$ eigenvalues of the Laplacians, compute the degree centrality over skeleta, and determine the all-pairs shortest path distance over $X^2_M$. The definitions of the chain matrix and Laplacians are provided in \citet{ribando2024combinatorial}. Intuitively, one can think of the chain matrix as consisting of many sampled formal sums over the $k$-cells. Effectively, this is somewhat analogous to the notion of collecting many paths over cells in a matrix. Each path is like a random walk over nodes in a graph, but instead of nodes, one has cells. The Laplacians effectively provide information about the connectedness of neighboring cells. One can compute statistics in this context and reduce dimensionality further. A technique one may apply is mean aggregation as in \citet{rahmani2021non}. All of these computed features are tensors that are concatenated and padded to uniform dimension. The resulting tensor is termed an ECC representation of molecule $M$. The algorithm is summarized in Figure \ref{fig:pactnet}.

\subsection{GNN Architecture}
\begin{figure}[t]
  \centering
  \includegraphics[width=0.95\linewidth]{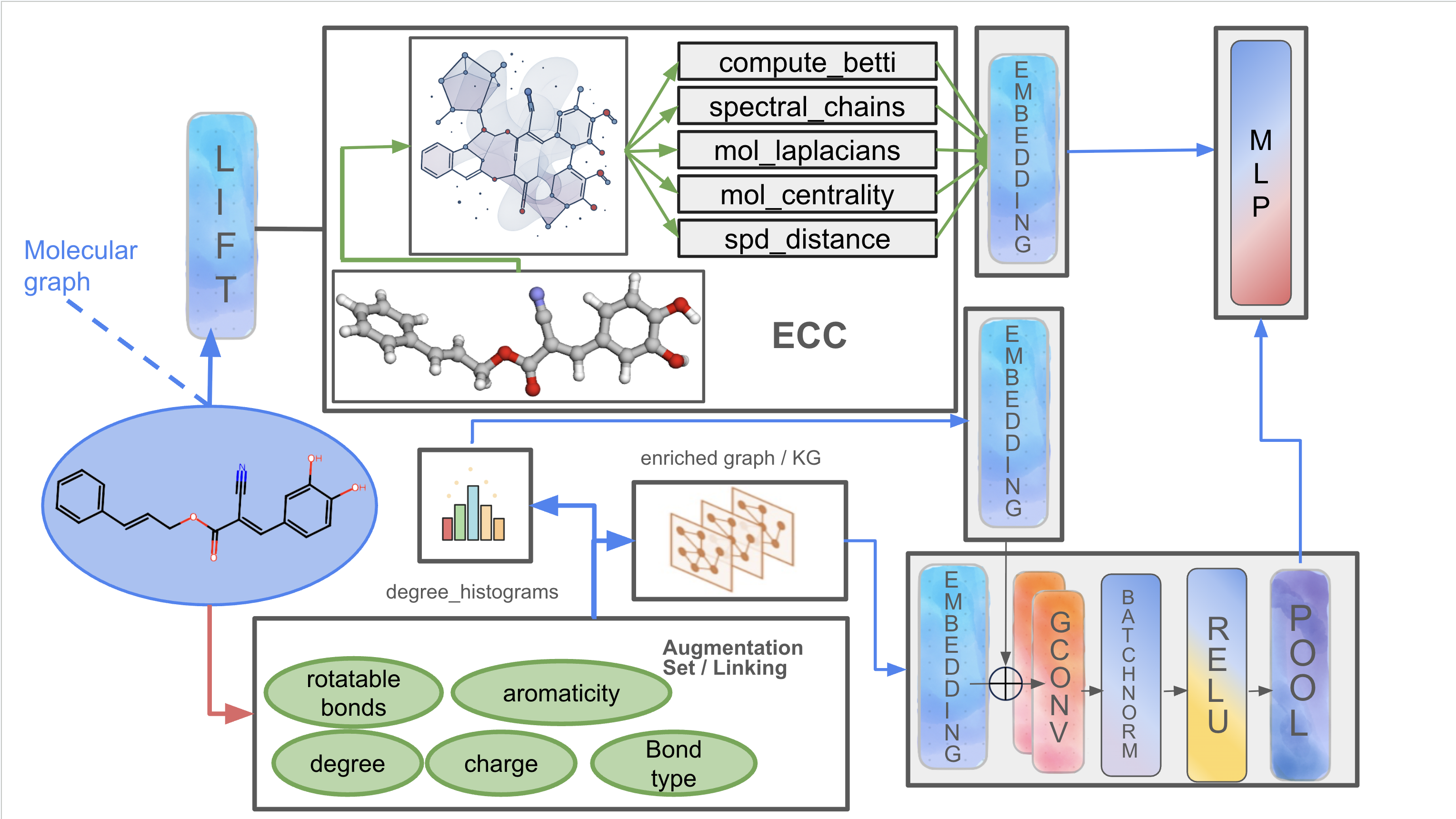}
  \caption{Overview of PACTNet architecture and ECC representation. The model takes a molecular graph as input, extracts spectral, chemical, and structural features, enriches the graph, and processes it through embedding and graph convolution layers before prediction.}
  \label{fig:pactnet}
\end{figure}

Upon determining the molecular graph $G_M$ for molecule $M$ we construct our ECC representation. Then simultaneously, we enrich our graph by adding in features related to rotatable bonds, aromaticity, degree, charge and bond type. Additionally, we compute the degree histograms and embed them. Afterward we can apply convolutional layers, batch norm and pooling. Our choice of convolution is the principal neighborhood aggregation scheme developed by \citet{corso2020principal}. The complete architecture is summarized in Figure \ref{fig:pactnet}.

\section{Experiments}

We run a wide array of experiments across standard benchmark datasets and provide performance metrics (RMSE/MAE) across these datasets. The datasets we use are summarized in table \ref{tab:datasets}. A deeper dive of these datasets can be found in Appendix \ref{appendix:data-deep-dive}.

\begin{table}[h!]
\centering
\caption{Overview of datasets used for experimental validation. All datasets are sourced from the MoleculeNet benchmark suite and involve predicting molecular properties.}
\label{tab:datasets}
\resizebox{\textwidth}{!}{
\begin{tabular}{@{}l l l l l@{}}
\toprule
\textbf{Dataset} & \textbf{Task Type} & \textbf{\# Molecules} & \textbf{Target Property} & \textbf{Scale/Domain} \\
\midrule
ESOL             & Regression         & 1,128                 & Water Solubility (log mol/L) & Biophysics (Small) \\
FreeSolv         & Regression         & 642                   & Hydration Free Energy (kcal/mol) & Biophysics (Small) \\
Lipophilicity    & Regression         & 4,200                 & Octanol/Water Distribution (logD) & Biophysics (Small) \\
Boiling Point    & Regression         & 2,983                 & Boiling Point (\textdegree C) & Physical Chemistry \\
QM9              & Regression         & 134k                  & Heat Capacity (Cv) & Quantum Mechanics \\
IC50             & Regression         & 2,822                 & pIC50 (-log10 of IC50) & Pharmacology \\
BindingDB        & Regression         & 4,614                 & Binding Affinity (Ki/Kd) & Pharmacology \\
\bottomrule
\end{tabular}}
\end{table}

\paragraph{Data Preprocessing} All datasets were subjected to a standard preprocessing pipeline. To manage computational load, we created subsets by taking a random sample of at most 2000 molecules from any dataset exceeding this size, and made these subsets publicly available for reproducibility. For the IC50 and BindingDB datasets, target values were transformed to pIC50, a standard practice in biochemical modeling \citep{kramer2012experimental}. For graph-based models (GNNs), molecular graphs were generated from SMILES strings using RDKit, following the protocol outlined by \citet{dablander2024investigating} and \citet{bento2020open}. Any molecules that failed parsing or contained null values were removed prior to experimentation.

\paragraph{Experimental Setup}
To estimate the generalization error of our entire learning procedure (including hyperparameter optimization), we employ a \textsc{5-fold nested cross-validation} scheme \citep{zhong2023nested}. A 5-fold partition was chosen to balance the trade-off between the bias and variance of the error estimate against the considerable computational expense of the nested procedure. The outer loop provides a nearly unbiased estimate of the true error, where each of the 5 folds serves as a hold-out validation set exactly once. The inner loop is used exclusively for hyperparameter selection on the remaining 4 folds.

Crucially, for all experiments, the outer-loop data splits were generated once using a fixed random seed (\texttt{random\_state=42}) and reused for every model and baseline. This ensures that any observed performance differences are attributable to the models themselves, not to variance in the data partitioning, thus satisfying the pairing assumption for subsequent statistical tests.

Within each inner loop of the nested cross-validation, we performed automated hyperparameter optimization (HPO) on an 80/20 split of the inner-loop data. We utilized a Tree-structured Parzen Estimator (TPE) for optimization, implemented in the Optuna library, chosen for its demonstrated efficiency over random search \citep{akiba2019optuna}.

The HPO was configured to run for 15 trials, with the objective of minimizing the validation set's Mean Absolute Error (MAE). For fairness, a consistent hyperparameter search space was used for all GNN models, defined as follows:
\begin{itemize}
    \item \textbf{Learning Rate ($\eta$):} Sampled from a log-uniform distribution over $[5 \times 10^{-4}, 1 \times 10^{-3}]$. This range was selected to ensure training stability, as preliminary experiments revealed that unconstrained searches led to highly erratic performance.
    \item \textbf{Hidden Dimensionality ($d_h$):} A categorical choice from $\{64, 128, 256\}$.
    \item \textbf{Batch Size ($B$):} A categorical choice from $\{32, 64\}$.
\end{itemize}

Individual models were trained using the Adam optimizer with default parameters ($\beta_1=0.9, \beta_2=0.999$) to minimize the Mean Absolute Error \citep{kingma2014adam}. Training was conducted for a maximum of 200 epochs. An early stopping criterion was implemented to mitigate overfitting to the inner-loop validation set. Specifically, we monitored the validation MAE and terminated training if no improvement was observed for a patience of 20 epochs. This patience value was chosen to be large enough to allow the model to escape shallow local minima without risking significant overfitting. Upon termination, the model parameters that yielded the lowest validation MAE during that run were restored for the evaluation on the outer-loop validation fold. The experimental setup is summarized in Algorithm \ref{alg:experiment} which can be found in Appendix \ref{appendix:experimental-algorithm}. Compute cost is discussed in Appendix \ref{appendix:compute}.


\section{Results}

In this section, we present and analyze the main experimental results. To maintain focus, we provide a detailed synthesis for a representative subset of the foundational datasets. These tasks highlight the key performance trends across the selected datasets. The comprehensive results for all seven datasets are provided in Appendix \ref{appendix:benchmark-tables}. We report that PactNet+ECC achieves the best validation set MAE, validation set RMSE, test set MAE, and test set RMSE across five of the seven selected datasets. For the other two datasets (BindingDB, IC50) PactNet+ECC test set performance is within $5\%$ of the best performing model. On the BindingDB dataset PactNet+ECC achieves the best validation RMSE and validation MAE. On the IC50 dataset PactNet+ECC achieves the best validation RMSE.

\subsection{Summary Table}

In this section, we provide the summary table for the QM9 benchmark dataset. Note that across all dataset benchmark tables we adopt the standard of reporting validation metrics as mean $\pm$ standard deviation from 5-fold nested cross-validation. The global test set metrics are reported as the mean $\pm$ the half-width of the 95\% confidence interval from a non-parametric bootstrap. The best validation set and test set RMSE and MAE for each dataset is highlighted in \textbf{bold}. We delve into the statistical analysis of these results in the next subsection. All seven summary tables are provided in Appendix \ref{appendix:benchmark-tables}. Moreover a deep dive into each dataset can be found in the Appendix \ref{appendix:data-deep-dive}.

\begin{table}[t]
\centering
\small 
\caption{Detailed performance of PACTNet on the QM9 dataset. Our model achieves the lowest validation and test RMSE and MAE, outperforming all other baselines. The units are in cal mol K.}
\label{tab:full_results}
\sisetup{separate-uncertainty, table-align-text-post=false}
\resizebox{\textwidth}{!}{\begin{tabular}{@{}ll S[table-format=2.3(4)] S[table-format=2.3(4)] S[table-format=2.4(4)] S[table-format=2.4(4)]@{}}
\toprule
\textbf{Dataset} & \textbf{Model (Rep.)} & {Val RMSE} & {Val MAE} & {Test RMSE} & {Test MAE} \\
\midrule
\multirow{13}{*}{QM9} 
 & \textbf{PACTNet (ECC)} & $\mathbf{0.999 \pm 0.099}$ & $\mathbf{0.659 \pm 0.069}$ & $\mathbf{1.0480 \pm 0.1805}$ & $\mathbf{0.6510 \pm 0.0815}$ \\
 \cmidrule(l){2-6}
 & GAT (ECFP) & 2.490 \pm 0.149 & 1.826 \pm 0.051 & 2.4370 \pm 0.2845 & 1.7870 \pm 0.1620 \\
 & GAT (SELFIES) & 1.898 \pm 0.072 & 1.493 \pm 0.048 & 1.8170 \pm 0.1360 & 1.4210 \pm 0.1135 \\
 & GAT (SMILES) & 1.896 \pm 0.071 & 1.488 \pm 0.052 & 1.7820 \pm 0.1350 & 1.4030 \pm 0.1105 \\
\cmidrule(l){2-6}
 & GCN (ECFP) & 2.483 \pm 0.171 & 1.811 \pm 0.085 & 2.4260 \pm 0.2380 & 1.8390 \pm 0.1615 \\
 & GCN (SELFIES) & 2.358 \pm 0.096 & 1.871 \pm 0.078 & 2.1550 \pm 0.1510 & 1.7270 \pm 0.1290 \\
 & GCN (SMILES) & 2.423 \pm 0.063 & 1.926 \pm 0.046 & 2.1260 \pm 0.1575 & 1.6980 \pm 0.1265 \\
 \cmidrule(l){2-6}
 & GIN (ECFP) & 2.583 \pm 0.113 & 1.913 \pm 0.066 & 2.5470 \pm 0.2480 & 1.9270 \pm 0.1660 \\
 & GIN (SELFIES) & 1.883 \pm 0.081 & 1.489 \pm 0.068 & 1.8120 \pm 0.1300 & 1.4350 \pm 0.1060 \\
 & GIN (SMILES) & 1.876 \pm 0.063 & 1.492 \pm 0.058 & 1.8120 \pm 0.1395 & 1.4190 \pm 0.1085 \\
  \cmidrule(l){2-6}
 & SAGE (ECFP) & 2.516 \pm 0.165 & 1.836 \pm 0.086 & 2.4860 \pm 0.2580 & 1.8310 \pm 0.1640 \\
 & SAGE (SELFIES) & 1.561 \pm 0.031 & 1.206 \pm 0.029 & 1.4880 \pm 0.1180 & 1.1560 \pm 0.0915 \\
 & SAGE (SMILES) & 1.552 \pm 0.050 & 1.209 \pm 0.034 & 1.4290 \pm 0.1120 & 1.0890 \pm 0.0905 \\
\bottomrule
\end{tabular}}
\end{table}

\subsection{Statistical Analysis}

To rigorously evaluate the performance of our proposed model, \textsc{PACTNet}, against existing baselines, we employ a robust statistical validation framework. Our primary evaluation metric is the Root Mean Squared Error (RMSE), obtained through a 5-fold nested cross-validation procedure as detailed in Algorithm~\ref{alg:experiment}. This nested CV provides a near unbiased estimate of each model's performance on the five outer folds, yielding a vector of five RMSE and MAE scores for each model being compared. To determine if the observed performance improvements of \textsc{PACTNet} are statistically significant, we conduct a multi-stage analysis. We justify our choices of our test in Appendix \ref{appendix:test-justify}.

\paragraph{Design}
Suppose we have a fixed prediction task and dataset $\mathcal{D}$. We partition our dataset into $D_{\mathrm{trainval}}$ and $D_{\mathrm{test}}$ as in Algorithm \ref{alg:experiment}. Then let $L$ be a loss function, either MAE or RMSE. Model development and HPO are confined to $D_{\mathrm{trainval}}$ where we perform nested  $K$-fold cross-validation as per the conventions in \citet{zhong2023nested}. After selection, the chosen model $M$ is refit with early stopping on all of $D_{\mathrm{trainval}}$ and evaluated once on $D_{\mathrm{test}}$. The reported values include a $95\%$ confidence interval determined via non-parametric bootstrap. Test-set metrics are reported descriptively and are not used for the formal within-dataset hypothesis tests. As discussed by \citet{BengioGrandvalet2004}, Nested Cross Validation mitigates selection bias in point estimation but does not render outer folds independent, which matters for valid inference \citep{VarmaSimon2006,BengioGrandvalet2004, cawley2010over}. However, it is still more rigorous than the alternative, namely vanilla Cross Validation \citep{VarmaSimon2006, cawley2010over}.

\paragraph{Estimand and hypotheses}
Fix a competitor $j$ and the designated control model $c$. Let $\ell^{(m)}_{i}$ denote the outer-fold validation loss on fold $i\in\{1,\dots,K\}$ for model $m\in\{c,j\}$.
Define the paired fold-wise contrast
\begin{equation}
  d^{(L)}_{i} \;\equiv\; \ell^{(j)}_{i} - \ell^{(c)}_{i},
  \qquad
  \bar d^{(L)} \;\equiv\; \frac{1}{K}\sum_{i=1}^{K} d^{(L)}_{i}
\end{equation}
as is convention; see \citet{NadeauBengio2003}, who define the per-split loss difference $L_{A\!-\!B}(j,i) = L_A(j,i) - L_B(j,i)$ and base inference on the mean of such differences across splits with a corrected variance, and \citet{Dietterich1998}.

By construction, \(d^{(L)}_{i}>0\) indicates the control attains smaller loss than the competitor on fold \(i\), and \(\bar d^{(L)}>0\) indicates an average advantage.
Our one-sided hypothesis for each loss \(L\) is
\begin{equation}
  H_{0}:\ \mathbb{E}\!\left[\bar d^{(L)}\right]\le 0
  \qquad \text{vs} \qquad
  H_{1}:\ \mathbb{E}\!\left[\bar d^{(L)}\right] > 0.
\end{equation}
This is the classic one-sided, one-sample (paired) t-test on the mean fold-wise difference \citep{bouckaert2004evaluating, NadeauBengio2003}. Classic constructions that justify the same paired-per-split paradigm include \citet{alpaydin1999combined} and \citet{Dietterich1998}.

\paragraph{Nadeau-Bengio corrected \(t\)-statistic (primary, within-dataset)}
To account for \(K\)-fold dependence we adopt the Nadeau-Bengio correction \citep{NadeauBengio2003}.
Let
\begin{equation}
  s^{2} \equiv \frac{1}{K-1}\sum_{i=1}^{K}\!\left(d^{(L)}_{i}-\bar d^{(L)}\right)^{2},
  \qquad
  \rho_{0} \equiv \frac{1}{K-1}.
\end{equation}
The corrected standard error and test statistic are
\begin{equation}
  \widehat{\mathrm{SE}}_{\mathrm{NB}}
  \equiv
  \sqrt{\Big(\tfrac{1}{K}+\rho_{0}\Big)\,s^{2}}
  =
  \sqrt{\left(\tfrac{1}{K}+\tfrac{1}{K-1}\right)s^{2}},
  \qquad
  t_{\mathrm{NB}} \equiv \frac{\bar d^{(L)}}{\widehat{\mathrm{SE}}_{\mathrm{NB}}}.
\end{equation}
We reference $t_{\mathrm{NB}}$ to a Student distribution with $\nu=K-1$ degrees of freedom and report the one-sided upper-tail $p$-value
\begin{equation}
  p = \Pr\left(T_{\nu} \ge t_{\mathrm{NB}}\,\right).
\end{equation}
Because the alternative (control superiority) is pre-specified, we avoid post-hoc tail selection.
A conservative NB-style interval for \(\bar d^{(L)}\) may be reported as
\begin{equation}
  \bar d^{(L)} \pm t_{1-\alpha/2,\nu}~\widehat{\mathrm{SE}}_{\mathrm{NB}}.
\end{equation}
The correction is intentionally conservative at small \(K\) (e.g., \(\rho_{0}=1/4\) when \(K=5\)) and effect sizes are considered.

\paragraph{Multiplicity}
Within a dataset the control is compared to $M-1$ competitors.
We control the family-wise error rate (FWER) at level $\alpha$ via Holm's sequentially rejective procedure \citep{Holm1979}, applied separately to the MAE family and to the RMSE (or MSE) family. Holm's method is uniformly more powerful than Bonferroni and provides strong FWER control without restrictive dependence assumptions.

\paragraph{Reporting}
For each competitor we report the one-sided \(p\)-value, and Holm-adjusted \(p\)-value for both MAE and RMSE. Additionally we provide the raw  mean difference in RMSE, $t_{NB-RMSE}$,  mean difference in MAE, $t_{NB-MAE}$,  and $95\%$ confidence intervals for NB RMSE and NB MAE. Note that with $K=5$ the degrees of freedom are small and the NB inflation is conservative. This is a deliberate choice that privileges calibrated size over power when the dependence structure is only partially observable \citep{BengioGrandvalet2004,NadeauBengio2003}. The resulting values from our statistical tests for all datasets are given in Appendix \ref{appendix:stats-summ-tables}.

\subsection{Results and Discussion}

\begin{table}[H]
\centering
\caption{Summary of statistical analysis across all datasets.}
\label{tab:statistical_summary}
\begin{tabular}{l l l l}
\toprule
\textbf{Dataset} & \textbf{\textit{p}-value ($<0.05$)} & \textbf{Holm \textit{p}-value ($<0.05$)}  & \textbf{Conclusion (PACTNet)} \\
\midrule
QM9 & \cmark & \cmark & Statistically Superior \\
\midrule
ESOL & \cmark & \cmark & Statistically Superior \\
\midrule
BOILINGPOINT & \cmark & \cmark & Statistically Superior \\
\midrule
LIPOPHIL & \cmark & \cmark & Statistically Superior \\
\midrule
FREESOLV & \cmark & \cmark & Statistically Superior\\
\midrule
BINDINGDB & \gminus  & \gminus & Statistical Tie\\
\midrule
IC50 & \gminus & \gminus & Statistical Tie \\
\bottomrule
\end{tabular}
\end{table}

\paragraph{Datasets: QM9, ESOL, BOILINGPOINT, LIPOPHIL, FREESOLV} Based on the analysis summarized in Table \ref{tab:statistical_summary}, our model, PACTNet, demonstrates statistically significant, superior performance. The observed $p$-values and Holm-corrected $p$-values are highly significant, leading to a confident rejection of the null hypothesis. This provides strong statistical evidence of PACTNet's superiority over the compared baseline models across these five datasets. Furthermore, the empirical results consistently show that PACTNet achieves the lowest Root Mean Squared Error (RMSE) and Mean Absolute Error (MAE) values on both the validation and test sets for every dataset (Appendix \ref{appendix:benchmark-tables}). This consistent and statistically significant improvement solidifies our conclusion that PACTNet is a superior model compared to the commonly used or gold-standard baselines.

\paragraph{Datasets: BINDINGDB, IC50}  As detailed in Table \ref{tab:statistical_summary}, PACTNet demonstrates performance statistically equivalent to the best-performing models on these two datasets. Although the statistical analysis did not find a significant difference that would allow us to confidently reject the null hypothesis, the empirical results are compelling. The differences in observed RMSE and MAE values across both validation and test sets are less than $5\%$. This negligible margin of error indicates that for practical applications, PACTNet's performance is indistinguishable from the top-performing baselines on these specific benchmarks. Therefore we determine the outcome is a statistical tie or perhaps negligibly worse.

\section{Conclusion}

Thus, we have shown that our approach delivers empirically and statistically significant improvements on most benchmarks. It strikes a promising balance, avoiding the computational cost of first-principles methods and the representational limits of purely 2D or string-based models, while retaining geometric expressivity and scalability. We have also introduced an algorithm for molecular embeddings that is efficient to compute, geometrically aware, and robust. Future work will focus on scaling our architecture to larger biomolecules, incorporating richer topological invariants or equivariant layers, and investigating the training dynamics of models built on our ECC embeddings. The main limitations of our study are its evaluation on a finite set of benchmarks with relatively small datasets, and the fact that our representation is not well suited for tasks requiring extremely fine-grained physical detail, such as quantum-level interaction modeling. These results strongly suggest that our framework can serve as a scalable and versatile foundation for the next generation of molecular machine learning models.

\newpage

\printbibliography

\newpage
\appendix

\section{Experimental Algorithm}
\label{appendix:experimental-algorithm}

\begin{algorithm}[H]
\caption{Experimental Design: Nested CV \& Final Evaluation}
\label{alg:experiment}

\KwIn{Dataset $D = (\mathcal{X}, \mathcal{Y})$, split into $D_{\text{trainval}}$ and $D_{\text{test}}$}
\KwIn{Number of outer folds $K=5$}
\KwOut{Unbiased validation metrics $(\mu_{\text{val}}, \sigma_{\text{val}})$}
\KwOut{Final test metrics with confidence intervals $\text{Metrics}_{\text{test}}$}
\BlankLine

\Comment{Part 1: Unbiased Performance Estimation}
$(\mu_{\text{val}}, \sigma_{\text{val}}) \leftarrow$ \FNestedCV{$D_{\text{trainval}}, K$}\;
\BlankLine

\Comment{Part 2: Final Model Training and Testing}
$\text{Metrics}_{\text{test}} \leftarrow$ \FFinalEval{$D_{\text{trainval}}, D_{\text{test}}$}\;
\BlankLine

\hrulefill
\BlankLine

\SetKwProg{Fn}{Function}{:}{}
\Fn{\FNestedCV{$D_{\text{trainval}}, K$}}{
    $\text{fold\_metrics} \leftarrow [\,]$\;
    \For{$k \leftarrow 1$ \KwTo $K$}{
        $(D_{\text{train}}^{(k)}, D_{\text{val}}^{(k)}) \leftarrow k\text{-th fold of } D_{\text{trainval}}$\;
        $\theta_k^* \leftarrow$ \FHPO{$D_{\text{train}}^{(k)}$} \Comment{Inner loop for HPO}
        $S_k \leftarrow \text{Scaler().fit}(D_{\text{train}}^{(k)})$\;
        $M_k \leftarrow$ \FTrain{$D_{\text{train}}^{(k)}, \theta_k^*, S_k$}\;
        $\text{metric}_k \leftarrow$ \FEval{$M_k, D_{\text{val}}^{(k)}, S_k$}\;
        Append $\text{metric}_k$ to $\text{fold\_metrics}$\;
    }
    \KwRet $(\text{mean}(\text{fold\_metrics}), \text{std}(\text{fold\_metrics}))$\;
}
\BlankLine

\Fn{\FFinalEval{$D_{\text{trainval}}, D_{\text{test}}$}}{
    $\theta_{\text{final}}^* \leftarrow$ \FHPO{$D_{\text{trainval}}$}\;
    $S_{\text{final}} \leftarrow \text{Scaler().fit}(D_{\text{trainval}})$\;
    $M_{\text{final}} \leftarrow$ \FTrain{$D_{\text{trainval}}, \theta_{\text{final}}^*, S_{\text{final}}$}\;
    $y_{\text{true}}, y_{\text{pred}} \leftarrow$ \FEval{$M_{\text{final}}, D_{\text{test}}, S_{\text{final}}$}\;
    $\text{metrics}_{\text{test}} \leftarrow$ \FBootstrap{$y_{\text{true}}, y_{\text{pred}}$}\;
    \KwRet $\text{metrics}_{\text{test}}$\;
}
\end{algorithm}

\section{Dataset Deep Dive}
\label{appendix:data-deep-dive}

These tasks were selected from the MoleculeNet, BindingDB, and QuanDB benchmark sets \citep{moleculenet, bindingdb, yang2024quandb}. MoleculeNet is a comprehensive and commonly used set of benchmark datasets for molecular property prediction. It was released as part of the DeepChem library and contains datasets from across quantum mechanics, physical chemistry, biophysics and physiology \citep{moleculenet}. The QM9 dataset is a comprehensive dataset which provides geometric, electronic and related thermodynamic properties. All molecules are modeled using density functional theory with the B3LYP functional and 6-31G basis set. The ESOL dataset is a small dataset consisting of water solubility data for compounds. Note that solubility is a property the molecule and not its conformers. FreeSolv provides experimentally determined hydration free energy of small molecules in water. Lipophilicity is an important feature of drug-like molecules that affects membrane permeability and solubility. The Lipophilicity dataset contains such molecules as well as their experimental octanol/water distribution. The BindingDB dataset contains experimentally determined protein-ligand binding affinities for numerous protein targets including isoforms and mutational variants \citet{bindingdb}. Affinity data across proteins is of key interest in computer-aided drug design for screening for drug candidates \citep{bindingdb}.  The QuanDB benchmark suite was designed for quantum chemical property prediction and materials discovery \citep{yang2024quandb}. It consists of a diverse array of organic  molecular compounds wherein each molecule is small namely having less than $40$ atoms. The Boiling Point dataset contains numerous molecules and their corresponding boiling point between $[-100 ^{\circ}C, 100 ^{\circ}C].$ The IC50 dataset contains small molecules and the concentration of those molecules which inhibits a biological process by $50\%$. This quantity is crucial for assessing bioactivity \cite{yang2024quandb}. This information is summarized along with the scale of each dataset and units in Table \ref{tab:datasets}.

\section{Statistical Test Justification}
\label{appendix:test-justify}

\paragraph{Justification of test choice}
Outer-fold validation losses under nested CV are (near) unbiased for the post-selection performance but are statistically dependent: training subsets overlap and validation folds partition the same finite sample. There is, in general, no universally unbiased estimator of the variance of $K$-fold CV \citep{BengioGrandvalet2004}. Consequently, fold-wise tests that ignore dependence such as Wilcoxon signed-rank, and classical paired $t$ do not achieve nominal size \citep{ramachandran2020mathematical}. Moreover, Wilcoxon also presumes symmetry of paired differences \citep{voraprateep2013robustness}. We deliberately avoid Friedman-type omnibus rank tests across folds, as these presume observations in different blocks are independent \citep{DemSar2006, eisinga2017exact}. This assumption does not apply to CV folds from a single dataset namely $D_{\mathrm{trainval}}$ \citep{DemSar2006}.

\section{Benchmark Tables}
\label{appendix:benchmark-tables}
\begin{table}[h]
\centering
\small 
\caption{Detailed performance of PACTNet on the QM9 dataset. Our model achieves the lowest validation and test RMSE and MAE, outperforming all other baselines. The units are in cal mol K.}
\label{tab:full_results}
\sisetup{separate-uncertainty, table-align-text-post=false}
\resizebox{\textwidth}{!}{\begin{tabular}{@{}ll S[table-format=2.3(4)] S[table-format=2.3(4)] S[table-format=2.4(4)] S[table-format=2.4(4)]@{}}
\toprule
\textbf{Dataset} & \textbf{Model (Rep.)} & {Val RMSE} & {Val MAE} & {Test RMSE} & {Test MAE} \\
\midrule
\multirow{13}{*}{QM9} 
 & \textbf{PACTNet (ECC)} & $\mathbf{0.999 \pm 0.099}$ & $\mathbf{0.659 \pm 0.069}$ & $\mathbf{1.0480 \pm 0.1805}$ & $\mathbf{0.6510 \pm 0.0815}$ \\
 \cmidrule(l){2-6}
 & GAT (ECFP) & 2.490 \pm 0.149 & 1.826 \pm 0.051 & 2.4370 \pm 0.2845 & 1.7870 \pm 0.1620 \\
 & GAT (SELFIES) & 1.898 \pm 0.072 & 1.493 \pm 0.048 & 1.8170 \pm 0.1360 & 1.4210 \pm 0.1135 \\
 & GAT (SMILES) & 1.896 \pm 0.071 & 1.488 \pm 0.052 & 1.7820 \pm 0.1350 & 1.4030 \pm 0.1105 \\
\cmidrule(l){2-6}
 & GCN (ECFP) & 2.483 \pm 0.171 & 1.811 \pm 0.085 & 2.4260 \pm 0.2380 & 1.8390 \pm 0.1615 \\
 & GCN (SELFIES) & 2.358 \pm 0.096 & 1.871 \pm 0.078 & 2.1550 \pm 0.1510 & 1.7270 \pm 0.1290 \\
 & GCN (SMILES) & 2.423 \pm 0.063 & 1.926 \pm 0.046 & 2.1260 \pm 0.1575 & 1.6980 \pm 0.1265 \\
 \cmidrule(l){2-6}
 & GIN (ECFP) & 2.583 \pm 0.113 & 1.913 \pm 0.066 & 2.5470 \pm 0.2480 & 1.9270 \pm 0.1660 \\
 & GIN (SELFIES) & 1.883 \pm 0.081 & 1.489 \pm 0.068 & 1.8120 \pm 0.1300 & 1.4350 \pm 0.1060 \\
 & GIN (SMILES) & 1.876 \pm 0.063 & 1.492 \pm 0.058 & 1.8120 \pm 0.1395 & 1.4190 \pm 0.1085 \\
  \cmidrule(l){2-6}
 & SAGE (ECFP) & 2.516 \pm 0.165 & 1.836 \pm 0.086 & 2.4860 \pm 0.2580 & 1.8310 \pm 0.1640 \\
 & SAGE (SELFIES) & 1.561 \pm 0.031 & 1.206 \pm 0.029 & 1.4880 \pm 0.1180 & 1.1560 \pm 0.0915 \\
 & SAGE (SMILES) & 1.552 \pm 0.050 & 1.209 \pm 0.034 & 1.4290 \pm 0.1120 & 1.0890 \pm 0.0905 \\
\bottomrule
\end{tabular}}
\end{table}

\begin{table}[h]
\centering
\small 
\caption{Detailed performance of PACTNet on the ESOL dataset. Our model achieves the lowest validation and test RMSE and MAE, outperforming all other baselines. The units are in $\log$ mol/L.}
\label{tab:full_results}
\sisetup{separate-uncertainty, table-align-text-post=false}
\resizebox{\textwidth}{!}{\begin{tabular}{@{}ll S[table-format=2.3(4)] S[table-format=2.3(4)] S[table-format=2.4(4)] S[table-format=2.4(4)]@{}}
\toprule
\textbf{Dataset} & \textbf{Model (Rep.)} & {Val RMSE} & {Val MAE} & {Test RMSE} & {Test MAE} \\
\midrule
\multirow{13}{*}{ESOL} 
& \textbf{PACTNet (ECC)} & $\mathbf{0.681 \pm 0.032}$ & $\mathbf{0.508 \pm 0.026}$ & $\mathbf{0.8290 \pm 0.1480}$ & $\mathbf{0.5930 \pm 0.0725}$ \\
\cmidrule(l){2-6}
& GAT (ECFP) & 1.226 \pm 0.070 & 0.929 \pm 0.051 & 1.1730 \pm 0.1300 & 0.8790 \pm 0.1060 \\
 & GAT (SELFIES) & 1.170 \pm 0.117 & 0.902 \pm 0.082 & 0.9850 \pm 0.1210 & 0.7440 \pm 0.0890 \\
 & GAT (SMILES) & 1.085 \pm 0.125 & 0.834 \pm 0.087 & 1.1400 \pm 0.1200 & 0.8710 \pm 0.0945 \\
\cmidrule(l){2-6}
 & GCN (ECFP) & 1.223 \pm 0.057 & 0.932 \pm 0.048 & 1.1710 \pm 0.1220 & 0.8880 \pm 0.0960 \\
 & GCN (SELFIES) & 1.279 \pm 0.114 & 0.977 \pm 0.085 & 1.2750 \pm 0.1530 & 0.9490 \pm 0.1070 \\
 & GCN (SMILES) & 1.240 \pm 0.172 & 0.958 \pm 0.125 & 1.2970 \pm 0.1690 & 0.9460 \pm 0.1125 \\
 \cmidrule(l){2-6}
 & GIN (ECFP) & 1.155 \pm 0.051 & 0.878 \pm 0.030 & 1.1070 \pm 0.1190 & 0.8500 \pm 0.0915 \\
 & GIN (SELFIES) & 1.247 \pm 0.172 & 0.940 \pm 0.138 & 1.3940 \pm 0.1890 & 0.9990 \pm 0.1235 \\
 & GIN (SMILES) & 1.196 \pm 0.082 & 0.908 \pm 0.068 & 1.3370 \pm 0.1570 & 0.9960 \pm 0.1170 \\
 \cmidrule(l){2-6}
 & SAGE (ECFP) & 1.218 \pm 0.076 & 0.922 \pm 0.058 & 1.1870 \pm 0.1175 & 0.8960 \pm 0.1040 \\
 & SAGE (SELFIES) & 1.055 \pm 0.116 & 0.802 \pm 0.080 & 0.9970 \pm 0.1215 & 0.7460 \pm 0.0840 \\
 & SAGE (SMILES) & 1.068 \pm 0.113 & 0.819 \pm 0.079 & 1.0890 \pm 0.1380 & 0.8140 \pm 0.0940 \\
\bottomrule
\end{tabular}
}
\end{table}

\begin{table}[H]
\centering
\small 
\caption{Detailed performance of PACTNet on the LIPOPHIL dataset. Our model achieves the lowest validation and test RMSE and MAE, outperforming all other baselines.}
\label{tab:full_results}
\sisetup{separate-uncertainty, table-align-text-post=false}
\resizebox{\textwidth}{!}{\begin{tabular}{@{}ll S[table-format=2.3(4)] S[table-format=2.3(4)] S[table-format=2.4(4)] S[table-format=2.4(4)]@{}}
\toprule
\textbf{Dataset} & \textbf{Model (Rep.)} & {Val RMSE} & {Val MAE} & {Test RMSE} & {Test MAE} \\
\midrule
\multirow{13}{*}{LIPOPHIL} 
 & \textbf{PACTNet (ECC)} & $\mathbf{0.717 \pm 0.027}$ & $\mathbf{0.531 \pm 0.023}$ & $\mathbf{0.7500 \pm 0.0505}$ & $\mathbf{0.5460 \pm 0.0335}$ \\
  \cmidrule(l){2-6}
 & GAT (ECFP) & 0.875 \pm 0.025 & 0.666 \pm 0.021 & 0.8630 \pm 0.0500 & 0.6580 \pm 0.0385 \\
 & GAT (SELFIES) & 1.037 \pm 0.052 & 0.833 \pm 0.044 & 1.0810 \pm 0.0460 & 0.8810 \pm 0.0415 \\
 & GAT (SMILES) & 1.031 \pm 0.044 & 0.825 \pm 0.034 & 1.0690 \pm 0.0505 & 0.8600 \pm 0.0445 \\
  \cmidrule(l){2-6}
 & GCN (ECFP) & 0.866 \pm 0.024 & 0.663 \pm 0.021 & 0.8380 \pm 0.0495 & 0.6380 \pm 0.0360 \\
 & GCN (SELFIES) & 1.070 \pm 0.041 & 0.863 \pm 0.033 & 1.1060 \pm 0.0460 & 0.9060 \pm 0.0420 \\
 & GCN (SMILES) & 1.075 \pm 0.045 & 0.869 \pm 0.033 & 1.1100 \pm 0.0470 & 0.9060 \pm 0.0450 \\
  \cmidrule(l){2-6}
 & GIN (ECFP) & 0.829 \pm 0.036 & 0.620 \pm 0.036 & 0.8080 \pm 0.0500 & 0.6050 \pm 0.0355 \\
 & GIN (SELFIES) & 1.062 \pm 0.065 & 0.848 \pm 0.054 & 1.0850 \pm 0.0530 & 0.8710 \pm 0.0430 \\
 & GIN (SMILES) & 1.072 \pm 0.043 & 0.862 \pm 0.036 & 1.0730 \pm 0.0495 & 0.8650 \pm 0.0410 \\
  \cmidrule(l){2-6}
 & SAGE (ECFP) & 0.865 \pm 0.021 & 0.663 \pm 0.022 & 0.8500 \pm 0.0480 & 0.6510 \pm 0.0365 \\
 & SAGE (SELFIES) & 0.947 \pm 0.038 & 0.750 \pm 0.027 & 1.0300 \pm 0.0615 & 0.8120 \pm 0.0435 \\
 & SAGE (SMILES) & 0.955 \pm 0.041 & 0.763 \pm 0.035 & 1.0160 \pm 0.0630 & 0.7970 \pm 0.0435 \\
\bottomrule
\end{tabular}
}
\end{table}

\begin{table}[H]
\centering
\small 
\caption{Detailed performance of PACTNet on the FREESOLV dataset. Our model achieves the lowest validation and test RMSE and MAE, outperforming all other baselines.}
\label{tab:full_results}
\sisetup{separate-uncertainty, table-align-text-post=false}
\resizebox{\textwidth}{!}{\begin{tabular}{@{}ll S[table-format=2.3(4)] S[table-format=2.3(4)] S[table-format=2.4(4)] S[table-format=2.4(4)]@{}}
\toprule
\textbf{Dataset} & \textbf{Model (Rep.)} & {Val RMSE} & {Val MAE} & {Test RMSE} & {Test MAE} \\
\midrule
\multirow{13}{*}{FREESOLV} 
 & \textbf{PACTNet (ECC)} & $\mathbf{1.313 \pm 0.111}$ & $\mathbf{0.857 \pm 0.064}$ & $\mathbf{1.4390 \pm 0.4430}$ & $\mathbf{0.8560 \pm 0.1925}$ \\
 \cmidrule(l){2-6}
 & GAT (ECFP) & 1.980 \pm 0.227 & 1.324 \pm 0.137 & 2.5360 \pm 0.7505 & 1.7260 \pm 0.3340 \\
 & GAT (SELFIES) & 2.787 \pm 0.361 & 2.059 \pm 0.334 & 3.6720 \pm 0.7910 & 2.7010 \pm 0.4400 \\
 & GAT (SMILES) & 2.777 \pm 0.373 & 2.070 \pm 0.305 & 3.7270 \pm 0.8130 & 2.7220 \pm 0.4425 \\
 \cmidrule(l){2-6}
 & GCN (ECFP) & 2.005 \pm 0.238 & 1.309 \pm 0.147 & 2.5380 \pm 0.8070 & 1.6090 \pm 0.3315 \\
 & GCN (SELFIES) & 3.486 \pm 0.216 & 2.546 \pm 0.097 & 3.7730 \pm 0.8485 & 2.7260 \pm 0.4650 \\
 & GCN (SMILES) & 3.381 \pm 0.240 & 2.486 \pm 0.184 & 3.8800 \pm 0.8570 & 2.8550 \pm 0.4605 \\
 \cmidrule(l){2-6}
 & GIN (ECFP) & 1.737 \pm 0.128 & 1.180 \pm 0.160 & 2.1720 \pm 0.5940 & 1.4270 \pm 0.2865 \\
 & GIN (SELFIES) & 3.427 \pm 0.170 & 2.553 \pm 0.133 & 3.8140 \pm 0.8350 & 2.7930 \pm 0.4465 \\
 & GIN (SMILES) & 3.454 \pm 0.199 & 2.528 \pm 0.123 & 3.6900 \pm 0.7775 & 2.6760 \pm 0.4545 \\
 \cmidrule(l){2-6}
 & SAGE (ECFP) & 1.894 \pm 0.162 & 1.285 \pm 0.150 & 2.3650 \pm 0.6605 & 1.5960 \pm 0.3140 \\
 & SAGE (SELFIES) & 2.498 \pm 0.429 & 1.851 \pm 0.389 & 3.7790 \pm 0.8365 & 2.7620 \pm 0.4305 \\
 & SAGE (SMILES) & 2.703 \pm 0.378 & 2.029 \pm 0.290 & 3.7890 \pm 0.8225 & 2.8020 \pm 0.4540 \\
\bottomrule
\end{tabular}
}
\end{table}

\begin{table}[H]
\centering
\small 
\caption{Detailed performance of PACTNet on the BOILINGPOINT dataset. Our model achieves the lowest validation and test RMSE and MAE, outperforming all other baselines.}
\label{tab:full_results}
\sisetup{separate-uncertainty, table-align-text-post=false}
\resizebox{\textwidth}{!}{\begin{tabular}{@{}ll S[table-format=2.3(4)] S[table-format=2.3(4)] S[table-format=2.4(4)] S[table-format=2.4(4)]@{}}
\toprule
\textbf{Dataset} & \textbf{Model (Rep.)} & {Val RMSE} & {Val MAE} & {Test RMSE} & {Test MAE} \\
\midrule
\multirow{13}{*}{BOILINGPT} 
 & \textbf{PACTNet (ECC)} & $\mathbf{49.089 \pm 2.425}$ & $\mathbf{33.893 \pm 1.592}$ & $\mathbf{50.6350 \pm 6.4695}$ & $\mathbf{33.2600 \pm 3.6895}$ \\
  \cmidrule(l){2-6}
 & GAT (ECFP) & 56.831 \pm 1.794 & 43.534 \pm 1.438 & 53.2660 \pm 4.5465 & 40.3050 \pm 3.2910 \\
 & GAT (SELFIES) & 55.602 \pm 1.883 & 42.244 \pm 1.894 & 56.0480 \pm 5.4735 & 40.4770 \pm 3.9400 \\
 & GAT (SMILES) & 55.240 \pm 2.739 & 41.337 \pm 2.469 & 60.7780 \pm 5.1350 & 45.3330 \pm 4.0420 \\
  \cmidrule(l){2-6}
 & GCN (ECFP) & 57.010 \pm 1.529 & 43.315 \pm 1.254 & 54.6140 \pm 4.5240 & 41.7490 \pm 3.6545 \\
 & GCN (SELFIES) & 61.200 \pm 1.848 & 47.294 \pm 1.035 & 62.3820 \pm 5.2935 & 46.8430 \pm 4.1720 \\
 & GCN (SMILES) & 61.586 \pm 1.292 & 47.662 \pm 1.277 & 59.0360 \pm 4.8840 & 43.9270 \pm 3.7900 \\
  \cmidrule(l){2-6}
 & GIN (ECFP) & 58.744 \pm 1.789 & 45.098 \pm 2.000 & 57.3840 \pm 5.0375 & 43.2210 \pm 3.6790 \\
 & GIN (SELFIES) & 58.629 \pm 0.867 & 44.500 \pm 0.975 & 57.8800 \pm 4.8570 & 42.4450 \pm 3.8065 \\
 & GIN (SMILES) & 59.282 \pm 2.271 & 45.235 \pm 2.034 & 57.1610 \pm 4.9560 & 42.1930 \pm 3.7810 \\
  \cmidrule(l){2-6}
 & SAGE (ECFP) & 57.161 \pm 1.253 & 43.714 \pm 0.873 & 55.2450 \pm 4.4265 & 42.1960 \pm 3.4290 \\
 & SAGE (SELFIES) & 54.345 \pm 2.264 & 40.948 \pm 1.797 & 53.0400 \pm 5.7445 & 37.4580 \pm 3.5905 \\
 & SAGE (SMILES) & 54.284 \pm 3.661 & 40.291 \pm 3.086 & 53.0110 \pm 5.5890 & 36.5630 \pm 3.7300 \\
\bottomrule
\end{tabular}
}
\end{table}

\begin{table}[H]
\centering
\small 
\caption{Detailed performance of PACTNet on the IC50 dataset. Our model achieves competitive validation and test RMSE and MAE, outperforming all other baselines.}
\label{tab:full_results}
\sisetup{separate-uncertainty, table-align-text-post=false}
\resizebox{\textwidth}{!}{\begin{tabular}{@{}ll S[table-format=2.3(4)] S[table-format=2.3(4)] S[table-format=2.4(4)] S[table-format=2.4(4)]@{}}
\toprule
\textbf{Dataset} & \textbf{Model (Rep.)} & {Val RMSE} & {Val MAE} & {Test RMSE} & {Test MAE} \\
\midrule
\multirow{13}{*}{IC50}
 & \textbf{PACTNet (ECC)} & $\mathbf{0.756 \pm 0.037}$ & 0.596 \pm 0.018 & 0.7500 \pm 0.0640 & 0.6060 \pm 0.0430 \\
  \cmidrule(l){2-6}
 & GAT (ECFP) & 0.768 \pm 0.038 & 0.596 \pm 0.016 & 0.7570 \pm 0.0730 & 0.5890 \pm 0.0475 \\
 & GAT (SELFIES) & 0.781 \pm 0.050 & 0.609 \pm 0.020 & 0.7400 \pm 0.0670 & 0.5880 \pm 0.0435 \\
 & GAT (SMILES) & 0.782 \pm 0.049 & 0.608 \pm 0.019 & 0.7410 \pm 0.0710 & 0.5810 \pm 0.0450 \\
  \cmidrule(l){2-6}
 & GCN (ECFP) & 0.764 \pm 0.036 & 0.596 \pm 0.019 & 0.7720 \pm 0.0735 & 0.5950 \pm 0.0505 \\
 & GCN (SELFIES) & 0.782 \pm 0.051 & 0.607 \pm 0.022 & 0.7450 \pm 0.0695 & 0.5890 \pm 0.0435 \\
 & GCN (SMILES) & 0.782 \pm 0.051 & 0.607 \pm 0.020 & 0.7410 \pm 0.0695 & 0.5860 \pm 0.0435 \\
  \cmidrule(l){2-6}
 & GIN (ECFP) & 0.782 \pm 0.035 & 0.605 \pm 0.013 & 0.7640 \pm 0.0735 & 0.5930 \pm 0.0485 \\
 & GIN (SELFIES) & 0.783 \pm 0.051 & 0.610 \pm 0.020 & 0.7410 \pm 0.0680 & 0.5900 \pm 0.0450 \\
 & GIN (SMILES) & 0.783 \pm 0.051 & 0.609 \pm 0.020 & 0.7400 \pm 0.0735 & 0.5860 \pm 0.0440 \\
  \cmidrule(l){2-6}
 & SAGE (ECFP) & 0.764 \pm 0.036 & $\mathbf{0.592 \pm 0.014}$ & 0.7840 \pm 0.0775 & 0.6040 \pm 0.0485 \\
 & SAGE (SELFIES) & 0.782 \pm 0.051 & 0.610 \pm 0.021 & $\mathbf{0.7360 \pm 0.0700}$ & 0.5840 \pm 0.0450 \\
 & SAGE (SMILES) & 0.782 \pm 0.050 & 0.608 \pm 0.020 & 0.7420 \pm 0.0735 & $\mathbf{0.5820 \pm 0.0445}$ \\
\bottomrule
\end{tabular}
}
\end{table}

\begin{table}[H]
\centering
\small 
\caption{Detailed performance of PACTNet on the BINDINGDB dataset. Our model achieves competitive validation and test RMSE and MAE, outperforming all other baselines.}
\label{tab:full_results}
\sisetup{separate-uncertainty, table-align-text-post=false}
\resizebox{\textwidth}{!}{\begin{tabular}{@{}ll S[table-format=2.3(4)] S[table-format=2.3(4)] S[table-format=2.4(4)] S[table-format=2.4(4)]@{}}
\toprule
\textbf{Dataset} & \textbf{Model (Rep.)} & {Val RMSE} & {Val MAE} & {Test RMSE} & {Test MAE} \\
\midrule
\multirow{13}{*}{BINDINGDB} 
 & \textbf{PACTNet (ECC)} & $\mathbf{1.479 \pm 0.034}$ & $\mathbf{1.211 \pm 0.030}$ & 1.7710 \pm 0.2420 & 1.3650 \pm 0.1080 \\
   \cmidrule(l){2-6}
 & GAT (ECFP) & 1.512 \pm 0.055 & 1.235 \pm 0.053 & 1.7750 \pm 0.2480 & 1.3360 \pm 0.1105 \\
 & GAT (SELFIES) & 1.488 \pm 0.029 & 1.227 \pm 0.023 & 1.7820 \pm 0.2440 & 1.3600 \pm 0.1145 \\
 & GAT (SMILES) & 1.491 \pm 0.028 & 1.230 \pm 0.020 & 1.7540 \pm 0.2395 & 1.3560 \pm 0.1100 \\
   \cmidrule(l){2-6}
 & GCN (ECFP) & 1.508 \pm 0.058 & 1.231 \pm 0.056 & 1.7620 \pm 0.2620 & 1.3220 \pm 0.1125 \\
 & GCN (SELFIES) & 1.492 \pm 0.031 & 1.231 \pm 0.023 & 1.7700 \pm 0.2485 & 1.3590 \pm 0.1110 \\
 & GCN (SMILES) & 1.495 \pm 0.031 & 1.234 \pm 0.023 & 1.8240 \pm 0.2750 & 1.3740 \pm 0.1210 \\
   \cmidrule(l){2-6}
 & GIN (ECFP) & 1.520 \pm 0.050 & 1.253 \pm 0.044 & 1.7830 \pm 0.2580 & 1.3540 \pm 0.1135 \\
 & GIN (SELFIES) & 1.494 \pm 0.030 & 1.231 \pm 0.025 & 1.7550 \pm 0.2410 & 1.3450 \pm 0.1095 \\
 & GIN (SMILES) & 1.494 \pm 0.032 & 1.230 \pm 0.027 & $\mathbf{1.7440 \pm 0.2420}$ & $\mathbf{1.3390 \pm 0.1070}$ \\
   \cmidrule(l){2-6}
 & SAGE (ECFP) & 1.506 \pm 0.054 & 1.235 \pm 0.046 & 1.7720 \pm 0.2610 & 1.3450 \pm 0.1140 \\
 & SAGE (SELFIES) & 1.493 \pm 0.031 & 1.234 \pm 0.024 & 1.9120 \pm 0.3785 & 1.3760 \pm 0.1220 \\
 & SAGE (SMILES) & 1.493 \pm 0.030 & 1.232 \pm 0.024 & 1.7870 \pm 0.2450 & 1.3810 \pm 0.1105 \\
\bottomrule
\end{tabular}
}
\end{table}

\section{Statistical Summary Tables}
\label{appendix:stats-summ-tables}

\subsection{QM9}

\begin{table}[H]
\centering
\small
\setlength{\tabcolsep}{4pt}
\caption{MAE: NB-corrected one-sided tests (outer folds, $K=5$) on dataset \texttt{qm9}; control \texttt{polyatomic\_polyatomic}.
Positive $\Delta$ (competitor $-$ control) favors control. Holm controls FWER.}
\label{tab:qm9_nb_compact_mae}
\begin{tabularx}{\linewidth}{@{}l X X X X X@{}}
\toprule
Comparison & {\(\Delta\)MAE} & {$t_{\text{NB}}$} & {CI\(_{\text{low}}\)} & {CI\(_{\text{high}}\)} & {p\(_{\text{Holm}}\)}\\
\midrule
polyatomic\_polyatomic vs gcn\_selfies & 1.212 & 48.842 & 1.143 & 1.281 & 6e-06 \\
polyatomic\_polyatomic vs gcn\_smiles & 1.267 & 25.966 & 1.132 & 1.403 & 7.2e-05 \\
polyatomic\_polyatomic vs gin\_ecfp & 1.254 & 22.974 & 1.103 & 1.406 & 0.000106 \\
polyatomic\_polyatomic vs gat\_smiles & 0.829 & 22.089 & 0.725 & 0.934 & 0.000112 \\
polyatomic\_polyatomic vs gat\_ecfp & 1.167 & 20.716 & 1.011 & 1.324 & 0.000125 \\
polyatomic\_polyatomic vs gin\_selfies & 0.831 & 20.866 & 0.720 & 0.941 & 0.000125 \\
polyatomic\_polyatomic vs gin\_smiles & 0.834 & 19.948 & 0.718 & 0.950 & 0.000125 \\
polyatomic\_polyatomic vs sage\_ecfp & 1.177 & 17.804 & 0.993 & 1.361 & 0.000146 \\
polyatomic\_polyatomic vs gcn\_ecfp & 1.153 & 15.862 & 0.951 & 1.354 & 0.000185 \\
polyatomic\_polyatomic vs gat\_selfies & 0.835 & 13.358 & 0.661 & 1.008 & 0.000272 \\
polyatomic\_polyatomic vs sage\_selfies & 0.548 & 9.634 & 0.390 & 0.706 & 0.000649 \\
polyatomic\_polyatomic vs sage\_smiles & 0.551 & 7.743 & 0.353 & 0.748 & 0.000749 \\
\bottomrule
\end{tabularx}
\end{table}

\begin{table}[H]
\centering
\small
\setlength{\tabcolsep}{4pt}
\caption{RMSE: NB-corrected one-sided tests (outer folds, $K=5$) on dataset \texttt{qm9}; control \texttt{polyatomic\_polyatomic}.
Positive $\Delta$ (competitor $-$ control) favors control. Holm controls FWER.}
\label{tab:qm9_nb_compact_rmse}
\begin{tabularx}{\linewidth}{@{}l X X X X X@{}}
\toprule
Comparison & {\(\Delta\)RMSE} & {$t_{\text{NB}}$} & {CI\(_{\text{low}}\)} & {CI\(_{\text{high}}\)} & {p\(_{\text{Holm}}\)}\\
\midrule
polyatomic\_polyatomic vs gcn\_smiles & 1.424 & 30.029 & 1.292 & 1.556 & 4.4e-05 \\
polyatomic\_polyatomic vs gcn\_selfies & 1.359 & 19.070 & 1.161 & 1.556 & 0.000245 \\
polyatomic\_polyatomic vs gin\_ecfp & 1.584 & 16.790 & 1.322 & 1.846 & 0.000369 \\
polyatomic\_polyatomic vs gat\_ecfp & 1.491 & 16.167 & 1.235 & 1.747 & 0.000385 \\
polyatomic\_polyatomic vs sage\_ecfp & 1.516 & 14.189 & 1.220 & 1.813 & 0.000573 \\
polyatomic\_polyatomic vs gcn\_ecfp & 1.484 & 12.955 & 1.166 & 1.802 & 0.000717 \\
polyatomic\_polyatomic vs gat\_smiles & 0.897 & 12.286 & 0.694 & 1.100 & 0.000756 \\
polyatomic\_polyatomic vs gin\_selfies & 0.884 & 11.020 & 0.661 & 1.107 & 0.000964 \\
polyatomic\_polyatomic vs gin\_smiles & 0.877 & 10.341 & 0.641 & 1.112 & 0.000987 \\
polyatomic\_polyatomic vs gat\_selfies & 0.899 & 8.979 & 0.621 & 1.177 & 0.00128 \\
polyatomic\_polyatomic vs sage\_selfies & 0.562 & 6.329 & 0.315 & 0.808 & 0.00319 \\
polyatomic\_polyatomic vs sage\_smiles & 0.553 & 6.171 & 0.304 & 0.802 & 0.00319 \\
\bottomrule
\end{tabularx}
\end{table}

\subsection{Lipophilicity}

\begin{table}[H]
\centering
\small
\setlength{\tabcolsep}{4pt}
\caption{MAE: NB-corrected one-sided tests (outer folds, $K=5$) on dataset \texttt{lipophil}; control \texttt{polyatomic\_polyatomic}.
Positive $\Delta$ (competitor $-$ control) favors control. Holm controls FWER.}
\label{tab:lipophil_nb_compact_mae}
\begin{tabularx}{\linewidth}{@{}l X X X X X@{}}
\toprule
Comparison & {\(\Delta\)MAE} & {$t_{\text{NB}}$} & {CI\(_{\text{low}}\)} & {CI\(_{\text{high}}\)} & {p\(_{\text{Holm}}\)}\\
\midrule
polyatomic\_polyatomic vs gcn\_selfies & 0.332 & 35.043 & 0.305 & 0.358 & 2.4e-05 \\
polyatomic\_polyatomic vs gcn\_smiles & 0.338 & 32.601 & 0.309 & 0.366 & 2.9e-05 \\
polyatomic\_polyatomic vs sage\_selfies & 0.219 & 15.687 & 0.180 & 0.257 & 0.000482 \\
polyatomic\_polyatomic vs gin\_smiles & 0.331 & 15.062 & 0.270 & 0.392 & 0.00051 \\
polyatomic\_polyatomic vs gat\_selfies & 0.301 & 14.563 & 0.244 & 0.359 & 0.000517 \\
polyatomic\_polyatomic vs sage\_smiles & 0.232 & 13.138 & 0.183 & 0.281 & 0.000678 \\
polyatomic\_polyatomic vs gat\_smiles & 0.293 & 11.414 & 0.222 & 0.365 & 0.00101 \\
polyatomic\_polyatomic vs gcn\_ecfp & 0.132 & 10.909 & 0.098 & 0.165 & 0.00101 \\
polyatomic\_polyatomic vs sage\_ecfp & 0.132 & 10.861 & 0.098 & 0.166 & 0.00101 \\
polyatomic\_polyatomic vs gat\_ecfp & 0.134 & 9.321 & 0.094 & 0.174 & 0.00111 \\
polyatomic\_polyatomic vs gin\_selfies & 0.317 & 7.518 & 0.200 & 0.434 & 0.00168 \\
polyatomic\_polyatomic vs gin\_ecfp & 0.089 & 3.640 & 0.021 & 0.157 & 0.011 \\
\bottomrule
\end{tabularx}
\end{table}

\begin{table}[H]
\centering
\small
\setlength{\tabcolsep}{4pt}
\caption{RMSE: NB-corrected one-sided tests (outer folds, $K=5$) on dataset \texttt{lipophil}; control \texttt{polyatomic\_polyatomic}.
Positive $\Delta$ (competitor $-$ control) favors control. Holm controls FWER.}
\label{tab:lipophil_nb_compact_rmse}
\begin{tabularx}{\linewidth}{@{}l X X X X X@{}}
\toprule
Comparison & {\(\Delta\)RMSE} & {$t_{\text{NB}}$} & {CI\(_{\text{low}}\)} & {CI\(_{\text{high}}\)} & {p\(_{\text{Holm}}\)}\\
\midrule
polyatomic\_polyatomic vs gcn\_selfies & 0.353 & 20.048 & 0.304 & 0.402 & 0.000219 \\
polyatomic\_polyatomic vs gcn\_smiles & 0.358 & 17.665 & 0.302 & 0.415 & 0.000332 \\
polyatomic\_polyatomic vs gin\_smiles & 0.355 & 13.628 & 0.283 & 0.427 & 0.000839 \\
polyatomic\_polyatomic vs gat\_selfies & 0.320 & 11.120 & 0.240 & 0.400 & 0.00167 \\
polyatomic\_polyatomic vs gat\_ecfp & 0.158 & 8.365 & 0.106 & 0.210 & 0.00259 \\
polyatomic\_polyatomic vs gat\_smiles & 0.314 & 8.565 & 0.212 & 0.416 & 0.00259 \\
polyatomic\_polyatomic vs gcn\_ecfp & 0.149 & 8.517 & 0.100 & 0.197 & 0.00259 \\
polyatomic\_polyatomic vs gin\_selfies & 0.345 & 6.855 & 0.205 & 0.485 & 0.00259 \\
polyatomic\_polyatomic vs sage\_ecfp & 0.148 & 9.091 & 0.103 & 0.193 & 0.00259 \\
polyatomic\_polyatomic vs sage\_selfies & 0.230 & 9.643 & 0.164 & 0.296 & 0.00259 \\
polyatomic\_polyatomic vs sage\_smiles & 0.238 & 9.620 & 0.170 & 0.307 & 0.00259 \\
polyatomic\_polyatomic vs gin\_ecfp & 0.112 & 4.077 & 0.036 & 0.189 & 0.00757 \\
\bottomrule
\end{tabularx}
\end{table}

\subsection{FreeSolv}

\begin{table}[H]
\centering
\small
\setlength{\tabcolsep}{4pt}
\caption{MAE: NB-corrected one-sided tests (outer folds, $K=5$) on dataset \texttt{freesolv}; control \texttt{polyatomic\_polyatomic}.
Positive $\Delta$ (competitor $-$ control) favors control. Holm controls FWER.}
\label{tab:freesolv_nb_compact_mae}
\begin{tabularx}{\linewidth}{@{}l X X X X X@{}}
\toprule
Comparison & {\(\Delta\)MAE} & {$t_{\text{NB}}$} & {CI\(_{\text{low}}\)} & {CI\(_{\text{high}}\)} & {p\(_{\text{Holm}}\)}\\
\midrule
polyatomic\_polyatomic vs gin\_smiles & 1.671 & 19.269 & 1.430 & 1.912 & 0.000256 \\
polyatomic\_polyatomic vs gcn\_selfies & 1.690 & 16.561 & 1.406 & 1.973 & 0.000428 \\
polyatomic\_polyatomic vs gin\_selfies & 1.697 & 15.350 & 1.390 & 2.004 & 0.000525 \\
polyatomic\_polyatomic vs gcn\_smiles & 1.629 & 11.909 & 1.249 & 2.009 & 0.00128 \\
polyatomic\_polyatomic vs gat\_ecfp & 0.467 & 7.425 & 0.292 & 0.641 & 0.00702 \\
polyatomic\_polyatomic vs sage\_smiles & 1.173 & 6.812 & 0.695 & 1.651 & 0.0085 \\
polyatomic\_polyatomic vs gat\_smiles & 1.213 & 6.224 & 0.672 & 1.754 & 0.00931 \\
polyatomic\_polyatomic vs gcn\_ecfp & 0.452 & 6.376 & 0.255 & 0.649 & 0.00931 \\
polyatomic\_polyatomic vs gat\_selfies & 1.202 & 5.427 & 0.587 & 1.817 & 0.0112 \\
polyatomic\_polyatomic vs sage\_ecfp & 0.428 & 4.617 & 0.171 & 0.686 & 0.0149 \\
polyatomic\_polyatomic vs sage\_selfies & 0.995 & 3.864 & 0.280 & 1.709 & 0.0181 \\
polyatomic\_polyatomic vs gin\_ecfp & 0.323 & 2.856 & 0.009 & 0.637 & 0.0231 \\
\bottomrule
\end{tabularx}
\end{table}

\begin{table}[H]
\centering
\small
\setlength{\tabcolsep}{4pt}
\caption{RMSE: NB-corrected one-sided tests (outer folds, $K=5$) on dataset \texttt{freesolv}; control \texttt{polyatomic\_polyatomic}.
Positive $\Delta$ (competitor $-$ control) favors control. Holm controls FWER.}
\label{tab:freesolv_nb_compact_rmse}
\begin{tabularx}{\linewidth}{@{}l X X X X X@{}}
\toprule
Comparison & {\(\Delta\)RMSE} & {$t_{\text{NB}}$} & {CI\(_{\text{low}}\)} & {CI\(_{\text{high}}\)} & {p\(_{\text{Holm}}\)}\\
\midrule
polyatomic\_polyatomic vs gin\_selfies & 2.113 & 11.592 & 1.607 & 2.619 & 0.0019 \\
polyatomic\_polyatomic vs gcn\_selfies & 2.172 & 11.171 & 1.632 & 2.712 & 0.00193 \\
polyatomic\_polyatomic vs gin\_smiles & 2.141 & 11.296 & 1.615 & 2.667 & 0.00193 \\
polyatomic\_polyatomic vs sage\_ecfp & 0.581 & 10.341 & 0.425 & 0.737 & 0.00222 \\
polyatomic\_polyatomic vs gcn\_smiles & 2.067 & 9.301 & 1.450 & 2.684 & 0.00297 \\
polyatomic\_polyatomic vs gat\_ecfp & 0.667 & 4.538 & 0.259 & 1.075 & 0.0146 \\
polyatomic\_polyatomic vs gat\_selfies & 1.474 & 5.882 & 0.778 & 2.169 & 0.0146 \\
polyatomic\_polyatomic vs gat\_smiles & 1.463 & 5.799 & 0.763 & 2.164 & 0.0146 \\
polyatomic\_polyatomic vs gcn\_ecfp & 0.691 & 5.112 & 0.316 & 1.067 & 0.0146 \\
polyatomic\_polyatomic vs gin\_ecfp & 0.424 & 5.376 & 0.205 & 0.643 & 0.0146 \\
polyatomic\_polyatomic vs sage\_selfies & 1.185 & 4.426 & 0.442 & 1.928 & 0.0146 \\
polyatomic\_polyatomic vs sage\_smiles & 1.390 & 5.690 & 0.712 & 2.068 & 0.0146 \\
\bottomrule
\end{tabularx}
\end{table}

\subsection{ESOL}

\begin{table}[H]
\centering
\small
\setlength{\tabcolsep}{4pt}
\caption{MAE: NB-corrected one-sided tests (outer folds, $K=5$) on dataset \texttt{esol}; control \texttt{polyatomic\_polyatomic}.
Positive $\Delta$ (competitor $-$ control) favors control. Holm controls FWER.}
\label{tab:esol_nb_compact_mae}
\begin{tabularx}{\linewidth}{@{}l X X X X X@{}}
\toprule
Comparison & {\(\Delta\)MAE} & {$t_{\text{NB}}$} & {CI\(_{\text{low}}\)} & {CI\(_{\text{high}}\)} & {p\(_{\text{Holm}}\)}\\
\midrule
polyatomic\_polyatomic vs gin\_ecfp & 0.370 & 46.410 & 0.348 & 0.392 & 8e-06 \\
polyatomic\_polyatomic vs gat\_ecfp & 0.421 & 17.654 & 0.355 & 0.487 & 0.000333 \\
polyatomic\_polyatomic vs gcn\_ecfp & 0.424 & 16.532 & 0.353 & 0.495 & 0.000392 \\
polyatomic\_polyatomic vs sage\_ecfp & 0.414 & 16.233 & 0.343 & 0.484 & 0.000392 \\
polyatomic\_polyatomic vs gat\_selfies & 0.393 & 7.609 & 0.250 & 0.537 & 0.00569 \\
polyatomic\_polyatomic vs gcn\_selfies & 0.469 & 7.753 & 0.301 & 0.637 & 0.00569 \\
polyatomic\_polyatomic vs gin\_smiles & 0.399 & 7.851 & 0.258 & 0.541 & 0.00569 \\
polyatomic\_polyatomic vs sage\_smiles & 0.310 & 7.347 & 0.193 & 0.427 & 0.00569 \\
polyatomic\_polyatomic vs gat\_smiles & 0.326 & 6.032 & 0.176 & 0.476 & 0.00762 \\
polyatomic\_polyatomic vs gcn\_smiles & 0.449 & 5.213 & 0.210 & 0.689 & 0.00969 \\
polyatomic\_polyatomic vs gin\_selfies & 0.431 & 4.552 & 0.168 & 0.694 & 0.00969 \\
polyatomic\_polyatomic vs sage\_selfies & 0.293 & 5.035 & 0.132 & 0.455 & 0.00969 \\
\bottomrule
\end{tabularx}
\end{table}

\begin{table}[H]
\centering
\small
\setlength{\tabcolsep}{4pt}
\caption{RMSE: NB-corrected one-sided tests (outer folds, $K=5$) on dataset \texttt{esol}; control \texttt{polyatomic\_polyatomic}.
Positive $\Delta$ (competitor $-$ control) favors control. Holm controls FWER.}
\label{tab:esol_nb_compact_rmse}
\begin{tabularx}{\linewidth}{@{}l X X X X X@{}}
\toprule
Comparison & {\(\Delta\)RMSE} & {$t_{\text{NB}}$} & {CI\(_{\text{low}}\)} & {CI\(_{\text{high}}\)} & {p\(_{\text{Holm}}\)}\\
\midrule
polyatomic\_polyatomic vs gin\_ecfp & 0.475 & 19.913 & 0.409 & 0.541 & 0.000225 \\
polyatomic\_polyatomic vs gcn\_ecfp & 0.543 & 17.259 & 0.456 & 0.630 & 0.000364 \\
polyatomic\_polyatomic vs gat\_ecfp & 0.545 & 14.954 & 0.444 & 0.646 & 0.000583 \\
polyatomic\_polyatomic vs sage\_ecfp & 0.538 & 13.238 & 0.425 & 0.650 & 0.000847 \\
polyatomic\_polyatomic vs gin\_smiles & 0.515 & 10.397 & 0.377 & 0.652 & 0.00193 \\
polyatomic\_polyatomic vs gcn\_selfies & 0.598 & 7.790 & 0.385 & 0.811 & 0.00513 \\
polyatomic\_polyatomic vs gat\_selfies & 0.490 & 6.728 & 0.288 & 0.692 & 0.00763 \\
polyatomic\_polyatomic vs sage\_smiles & 0.388 & 5.838 & 0.203 & 0.572 & 0.0107 \\
polyatomic\_polyatomic vs gat\_smiles & 0.404 & 5.405 & 0.196 & 0.612 & 0.0113 \\
polyatomic\_polyatomic vs gcn\_smiles & 0.559 & 5.047 & 0.252 & 0.867 & 0.0113 \\
polyatomic\_polyatomic vs gin\_selfies & 0.566 & 4.778 & 0.237 & 0.896 & 0.0113 \\
polyatomic\_polyatomic vs sage\_selfies & 0.374 & 4.407 & 0.138 & 0.609 & 0.0113 \\
\bottomrule
\end{tabularx}
\end{table}

\subsection{BoilingPoint}

\begin{table}[H]
\centering
\small
\setlength{\tabcolsep}{4pt}
\caption{MAE: NB-corrected one-sided tests (outer folds, $K=5$) on dataset \texttt{boilingpoint}; control \texttt{polyatomic\_polyatomic}.
Positive $\Delta$ (competitor $-$ control) favors control. Holm controls FWER.}
\centering
\label{tab:boilingpoint_nb_compact_mae}
\begin{tabularx}{\linewidth}{@{}l X X X X X@{}}
\toprule
Comparison & {\(\Delta\)MAE} & {$t_{\text{NB}}$} & {CI\(_{\text{low}}\)} & {CI\(_{\text{high}}\)} & {p\(_{\text{Holm}}\)}\\
\midrule
polyatomic\_polyatomic vs gcn\_selfies & 13.401 & 9.426 & 9.454 & 17.348 & 0.00424 \\
polyatomic\_polyatomic vs gcn\_smiles & 13.768 & 6.842 & 8.181 & 19.356 & 0.0131 \\
polyatomic\_polyatomic vs gin\_ecfp & 11.204 & 6.414 & 6.355 & 16.054 & 0.0131 \\
polyatomic\_polyatomic vs gin\_selfies & 10.606 & 6.218 & 5.871 & 15.342 & 0.0131 \\
polyatomic\_polyatomic vs gin\_smiles & 11.342 & 6.554 & 6.537 & 16.147 & 0.0131 \\
polyatomic\_polyatomic vs sage\_ecfp & 9.821 & 6.713 & 5.759 & 13.883 & 0.0131 \\
polyatomic\_polyatomic vs gcn\_ecfp & 9.421 & 5.680 & 4.816 & 14.027 & 0.0142 \\
polyatomic\_polyatomic vs gat\_ecfp & 9.641 & 5.181 & 4.474 & 14.807 & 0.0165 \\
polyatomic\_polyatomic vs sage\_selfies & 7.055 & 4.786 & 2.962 & 11.147 & 0.0175 \\
polyatomic\_polyatomic vs gat\_selfies & 8.351 & 3.356 & 1.441 & 15.260 & 0.0426 \\
polyatomic\_polyatomic vs gat\_smiles & 7.444 & 3.000 & 0.554 & 14.335 & 0.0426 \\
polyatomic\_polyatomic vs sage\_smiles & 6.398 & 2.356 & -1.141 & 13.937 & 0.0426\\
\bottomrule
\end{tabularx}
\end{table}

\begin{table}[H]
\centering
\small
\setlength{\tabcolsep}{4pt}
\caption{RMSE: NB-corrected one-sided tests (outer folds, $K=5$) on dataset \texttt{boilingpoint}; control \texttt{polyatomic\_polyatomic}.
Positive $\Delta$ (competitor $-$ control) favors control. Holm controls FWER.}
\label{tab:boilingpoint_nb_compact_rmse}
\centering
\begin{tabularx}{\linewidth}{@{}l X X X X X@{}}
\toprule
Comparison & {\(\Delta\)RMSE} & {$t_{\text{NB}}$} & {CI\(_{\text{low}}\)} & {CI\(_{\text{high}}\)} & {p\(_{\text{Holm}}\)}\\
\midrule
polyatomic\_polyatomic vs gcn\_selfies & 12.111 & 7.335 & 7.527 & 16.695 & 0.011 \\
polyatomic\_polyatomic vs gcn\_smiles & 12.497 & 6.176 & 6.879 & 18.114 & 0.0192 \\
polyatomic\_polyatomic vs gin\_ecfp & 9.655 & 5.861 & 5.081 & 14.229 & 0.0211 \\
polyatomic\_polyatomic vs sage\_ecfp & 8.072 & 5.138 & 3.710 & 12.433 & 0.0306 \\
polyatomic\_polyatomic vs gcn\_ecfp & 7.921 & 4.721 & 3.263 & 12.579 & 0.0366 \\
polyatomic\_polyatomic vs gat\_ecfp & 7.742 & 4.407 & 2.865 & 12.619 & 0.0407 \\
polyatomic\_polyatomic vs gin\_selfies & 9.540 & 4.336 & 3.431 & 15.649 & 0.0407 \\
polyatomic\_polyatomic vs gin\_smiles & 10.193 & 3.937 & 3.004 & 17.381 & 0.0425 \\
polyatomic\_polyatomic vs gat\_selfies & 6.513 & 3.322 & 1.069 & 11.956 & 0.0587 \\
polyatomic\_polyatomic vs gat\_smiles & 6.151 & 2.813 & 0.081 & 12.222 & 0.0722 \\
polyatomic\_polyatomic vs sage\_selfies & 5.256 & 2.449 & -0.703 & 11.215 & 0.0722 \\
polyatomic\_polyatomic vs sage\_smiles & 5.195 & 1.562 & -4.040 & 14.429 & 0.0967 \\
\bottomrule
\end{tabularx}
\end{table}

\subsection{BindingDB}

\begin{table}[H]
\centering
\small
\setlength{\tabcolsep}{4pt}
\caption{MAE: NB-corrected one-sided tests (outer folds, $K=5$) on dataset \texttt{bindingdb}; control \texttt{polyatomic\_polyatomic}.
Positive $\Delta$ (competitor $-$ control) favors control. Holm controls FWER.}
\label{tab:bindingdb_nb_compact_mae}
\centering
\begin{tabularx}{\linewidth}{@{}l X X X X X@{}}
\toprule
Comparison & {\(\Delta\)MAE} & {$t_{\text{NB}}$} & {CI\(_{\text{low}}\)} & {CI\(_{\text{high}}\)} & {p\(_{\text{Holm}}\)}\\
\midrule
polyatomic\_polyatomic vs sage\_smiles & 0.021 & 3.758 & 0.005 & 0.036 & 0.119 \\
polyatomic\_polyatomic vs sage\_selfies & 0.022 & 3.236 & 0.003 & 0.041 & 0.175 \\
polyatomic\_polyatomic vs gat\_selfies & 0.015 & 2.085 & -0.005 & 0.035 & 0.386 \\
polyatomic\_polyatomic vs gcn\_selfies & 0.020 & 1.929 & -0.009 & 0.049 & 0.386 \\
polyatomic\_polyatomic vs gcn\_smiles & 0.023 & 2.334 & -0.004 & 0.050 & 0.386 \\
polyatomic\_polyatomic vs gin\_selfies & 0.020 & 2.365 & -0.003 & 0.043 & 0.386 \\
polyatomic\_polyatomic vs gin\_smiles & 0.019 & 2.214 & -0.005 & 0.042 & 0.386 \\
polyatomic\_polyatomic vs gat\_smiles & 0.018 & 1.546 & -0.015 & 0.051 & 0.425 \\
polyatomic\_polyatomic vs gin\_ecfp & 0.042 & 1.671 & -0.028 & 0.112 & 0.425 \\
polyatomic\_polyatomic vs gat\_ecfp & 0.023 & 0.918 & -0.048 & 0.094 & 0.431 \\
polyatomic\_polyatomic vs gcn\_ecfp & 0.019 & 0.743 & -0.053 & 0.092 & 0.431 \\
polyatomic\_polyatomic vs sage\_ecfp & 0.024 & 1.227 & -0.030 & 0.077 & 0.431 \\
\bottomrule
\end{tabularx}
\end{table}

\begin{table}[H]
\centering
\small
\setlength{\tabcolsep}{4pt}
\caption{RMSE: NB-corrected one-sided tests (outer folds, $K=5$) on dataset \texttt{bindingdb}; control \texttt{polyatomic\_polyatomic}.
Positive $\Delta$ (competitor $-$ control) favors control. Holm controls FWER.}
\label{tab:bindingdb_nb_compact_rmse}
\centering
\begin{tabularx}{\linewidth}{@{}l X X X X X@{}}
\toprule
Comparison & {\(\Delta\)RMSE} & {$t_{\text{NB}}$} & {CI\(_{\text{low}}\)} & {CI\(_{\text{high}}\)} & {p\(_{\text{Holm}}\)}\\
\midrule
polyatomic\_polyatomic vs gat\_ecfp & 0.033 & 1.563 & -0.026 & 0.093 & 0.825 \\
polyatomic\_polyatomic vs gat\_selfies & 0.009 & 0.737 & -0.025 & 0.043 & 0.825 \\
polyatomic\_polyatomic vs gat\_smiles & 0.013 & 0.937 & -0.025 & 0.050 & 0.825 \\
polyatomic\_polyatomic vs gcn\_ecfp & 0.029 & 1.386 & -0.030 & 0.088 & 0.825 \\
polyatomic\_polyatomic vs gcn\_selfies & 0.013 & 1.029 & -0.023 & 0.049 & 0.825 \\
polyatomic\_polyatomic vs gcn\_smiles & 0.016 & 1.282 & -0.018 & 0.050 & 0.825 \\
polyatomic\_polyatomic vs gin\_ecfp & 0.041 & 1.853 & -0.021 & 0.103 & 0.825 \\
polyatomic\_polyatomic vs gin\_selfies & 0.016 & 1.775 & -0.009 & 0.040 & 0.825 \\
polyatomic\_polyatomic vs gin\_smiles & 0.015 & 1.620 & -0.011 & 0.042 & 0.825 \\
polyatomic\_polyatomic vs sage\_ecfp & 0.027 & 1.377 & -0.027 & 0.081 & 0.825 \\
polyatomic\_polyatomic vs sage\_selfies & 0.015 & 1.480 & -0.013 & 0.042 & 0.825 \\
polyatomic\_polyatomic vs sage\_smiles & 0.014 & 1.785 & -0.008 & 0.037 & 0.825 \\
\bottomrule
\end{tabularx}
\end{table}

\subsection{IC50}

\begin{table}[H]
\centering
\small
\setlength{\tabcolsep}{4pt}
\caption{MAE: NB-corrected one-sided tests (outer folds, $K=5$) on dataset \texttt{ic50}; control \texttt{polyatomic\_polyatomic}.
Positive $\Delta$ (competitor $-$ control) favors control. Holm controls FWER.}
\label{tab:ic50_nb_compact_mae}
\begin{tabularx}{\linewidth}{@{}l X X X X X@{}}
\toprule
Comparison & {\(\Delta\)MAE} & {$t_{\text{NB}}$} & {CI\(_{\text{low}}\)} & {CI\(_{\text{high}}\)} & {p\(_{\text{Holm}}\)}\\
\midrule
polyatomic\_polyatomic vs gat\_ecfp & -0.000 & -0.023 & -0.037 & 0.036 & 1 \\
polyatomic\_polyatomic vs gat\_selfies & 0.013 & 1.179 & -0.017 & 0.043 & 1 \\
polyatomic\_polyatomic vs gat\_smiles & 0.012 & 1.159 & -0.017 & 0.041 & 1 \\
polyatomic\_polyatomic vs gcn\_ecfp & -0.000 & -0.008 & -0.023 & 0.023 & 1 \\
polyatomic\_polyatomic vs gcn\_selfies & 0.011 & 0.840 & -0.024 & 0.046 & 1 \\
polyatomic\_polyatomic vs gcn\_smiles & 0.011 & 1.028 & -0.019 & 0.041 & 1 \\
polyatomic\_polyatomic vs gin\_ecfp & 0.009 & 0.825 & -0.021 & 0.038 & 1 \\
polyatomic\_polyatomic vs gin\_selfies & 0.014 & 1.507 & -0.012 & 0.039 & 1 \\
polyatomic\_polyatomic vs gin\_smiles & 0.013 & 1.296 & -0.014 & 0.040 & 1 \\
polyatomic\_polyatomic vs sage\_ecfp & -0.004 & -0.463 & -0.030 & 0.021 & 1 \\
polyatomic\_polyatomic vs sage\_selfies & 0.014 & 1.206 & -0.018 & 0.045 & 1 \\
polyatomic\_polyatomic vs sage\_smiles & 0.012 & 1.050 & -0.020 & 0.043 & 1 \\
\bottomrule
\end{tabularx}
\end{table}

\begin{table}[H]
\centering
\small
\setlength{\tabcolsep}{4pt}
\caption{RMSE: NB-corrected one-sided tests (outer folds, $K=5$) on dataset \texttt{ic50}; control \texttt{polyatomic\_polyatomic}.
Positive $\Delta$ (competitor $-$ control) favors control. Holm controls FWER.}
\label{tab:ic50_nb_compact_rmse}
\begin{tabularx}{\linewidth}{@{}l X X X X X@{}}
\toprule
Comparison & {\(\Delta\)RMSE} & {$t_{\text{NB}}$} & {CI\(_{\text{low}}\)} & {CI\(_{\text{high}}\)} & {p\(_{\text{Holm}}\)}\\
\midrule
polyatomic\_polyatomic vs gin\_ecfp & 0.026 & 5.608 & 0.013 & 0.039 & 0.0298 \\
polyatomic\_polyatomic vs gat\_ecfp & 0.012 & 1.099 & -0.018 & 0.042 & 1 \\
polyatomic\_polyatomic vs gat\_selfies & 0.025 & 1.203 & -0.032 & 0.081 & 1 \\
polyatomic\_polyatomic vs gat\_smiles & 0.025 & 1.238 & -0.031 & 0.082 & 1 \\
polyatomic\_polyatomic vs gcn\_ecfp & 0.007 & 0.693 & -0.022 & 0.037 & 1 \\
polyatomic\_polyatomic vs gcn\_selfies & 0.026 & 1.277 & -0.031 & 0.083 & 1 \\
polyatomic\_polyatomic vs gcn\_smiles & 0.026 & 1.282 & -0.030 & 0.082 & 1 \\
polyatomic\_polyatomic vs gin\_selfies & 0.027 & 1.374 & -0.027 & 0.081 & 1 \\
polyatomic\_polyatomic vs gin\_smiles & 0.027 & 1.338 & -0.029 & 0.082 & 1 \\
polyatomic\_polyatomic vs sage\_ecfp & 0.007 & 0.789 & -0.019 & 0.034 & 1 \\
polyatomic\_polyatomic vs sage\_selfies & 0.026 & 1.239 & -0.032 & 0.085 & 1 \\
polyatomic\_polyatomic vs sage\_smiles & 0.026 & 1.244 & -0.031 & 0.082 & 1 \\
\bottomrule
\end{tabularx}
\end{table}

\section{Compute Cost}
\label{appendix:compute}
All experiments were performed using an AWS \textsc{m8g.4xlarge} (CPU) instance.


\newpage
\section*{NeurIPS Paper Checklist}

\begin{enumerate}

\item {\bf Claims}
    \item[] Question: Do the main claims made in the abstract and introduction accurately reflect the paper's contributions and scope?
    \item[] Answer: \answerYes{} 
    \item[] Justification: We provide all statistical evidence and rigorous experimental design to demonstrate all performance based claims. Claims of topological feature integration are justified and provided in the methods section. Our learning method is computationally efficient and our GNN is novel.

\item {\bf Limitations}
    \item[] Question: Does the paper discuss the limitations of the work performed by the authors?
    \item[] Answer: \answerYes{} 
    \item[] Justification: See discussion.

\item {\bf Theory assumptions and proofs}
    \item[] Question: For each theoretical result, does the paper provide the full set of assumptions and a complete (and correct) proof?
    \item[] Answer: \answerNA{} 
    \item[] Justification: No explicit proofs provided.

    \item {\bf Experimental result reproducibility}
    \item[] Question: Does the paper fully disclose all the information needed to reproduce the main experimental results of the paper to the extent that it affects the main claims and/or conclusions of the paper (regardless of whether the code and data are provided or not)?
    \item[] Answer: \answerYes{} 
    \item[] Justification: Full experimental algorithm provided in the appendix and discussed extensively in the main experiment section. Additionally, all code is open source.

\item {\bf Open access to data and code}
    \item[] Question: Does the paper provide open access to the data and code, with sufficient instructions to faithfully reproduce the main experimental results, as described in supplemental material?
    \item[] Answer: \answerYes{}
    \item[] Justification: Open sourced code and all datasets. See github.

\item {\bf Experimental setting/details}
    \item[] Question: Does the paper specify all the training and test details (e.g., data splits, hyperparameters, how they were chosen, type of optimizer, etc.) necessary to understand the results?
    \item[] Answer: \answerYes{} 
    \item[] Justification: See experiment section.

\item {\bf Experiment statistical significance}
    \item[] Question: Does the paper report error bars suitably and correctly defined or other appropriate information about the statistical significance of the experiments?
    \item[] Answer: \answerYes{} 
    \item[] Justification: Absolutely, that and more see statistical analysis section of results.

\item {\bf Experiments compute resources}
    \item[] Question: For each experiment, does the paper provide sufficient information on the computer resources (type of compute workers, memory, time of execution) needed to reproduce the experiments?
    \item[] Answer: \answerYes{} 
    \item[] Justification: In the appendix yes.
    
\item {\bf Code of ethics}
    \item[] Question: Does the research conducted in the paper conform, in every respect, with the NeurIPS Code of Ethics \url{https://neurips.cc/public/EthicsGuidelines}?
    \item[] Answer: \answerYes{} 
    \item[] Justification: We were ethical undoubtedly and have read the guidelines.

\item {\bf Broader impacts}
    \item[] Question: Does the paper discuss both potential positive societal impacts and negative societal impacts of the work performed?
    \item[] Answer: \answerNA{} 
    \item[] Justification: No societal impact of the work performed.
    
\item {\bf Safeguards}
    \item[] Question: Does the paper describe safeguards that have been put in place for responsible release of data or models that have a high risk for misuse (e.g., pretrained language models, image generators, or scraped datasets)?
    \item[] Answer: \answerNA{} 
    \item[] Justification: No such risk posed.

\item {\bf Licenses for existing assets}
    \item[] Question: Are the creators or original owners of assets (e.g., code, data, models), used in the paper, properly credited and are the license and terms of use explicitly mentioned and properly respected?
    \item[] Answer: \answerYes{} 
    \item[] Justification: Absolutely.

\item {\bf New assets}
    \item[] Question: Are new assets introduced in the paper well documented and is the documentation provided alongside the assets?
    \item[] Answer: \answerYes{} 
    \item[] Justification: Code is open source and clearly documented with readme files.

\item {\bf Crowdsourcing and research with human subjects}
    \item[] Question: For crowdsourcing experiments and research with human subjects, does the paper include the full text of instructions given to participants and screenshots, if applicable, as well as details about compensation (if any)? 
    \item[] Answer: \answerNA{} 
    \item[] Justification: No human subjects.

\item {\bf Institutional review board (IRB) approvals or equivalent for research with human subjects}
    \item[] Question: Does the paper describe potential risks incurred by study participants, whether such risks were disclosed to the subjects, and whether Institutional Review Board (IRB) approvals (or an equivalent approval/review based on the requirements of your country or institution) were obtained?
    \item[] Answer: \answerNA{} 
    \item[] Justification: No human subjects.

\item {\bf Declaration of LLM usage}
    \item[] Question: Does the paper describe the usage of LLMs if it is an important, original, or non-standard component of the core methods in this research? Note that if the LLM is used only for writing, editing, or formatting purposes and does not impact the core methodology, scientific rigorousness, or originality of the research, declaration is not required.
    \item[] Answer: \answerNA{} 
    \item[] Justification: Research does not involve LLMs in any way/shape/form.

\end{enumerate}

\end{document}